\documentclass[10pt,twocolumn,letterpaper]{article}

\usepackage[pagenumbers]{cvpr} 

\definecolor{cvprblue}{rgb}{0.21,0.49,0.74}
\usepackage[pagebackref,breaklinks,colorlinks,allcolors=cvprblue]{hyperref}
\usepackage{hyperref}       
\usepackage{url}            
\usepackage{booktabs}       
\usepackage{amsfonts}       
\usepackage{nicefrac}       
\usepackage{microtype}      
\usepackage{xcolor}         
\usepackage{graphicx} 
\usepackage{amsmath}
\usepackage{multirow}  
\usepackage{caption}
\usepackage{algorithm}
\usepackage{algpseudocode}
\usepackage{subcaption}  
\usepackage{siunitx}  
\usepackage{adjustbox}  
\usepackage{tabularx}   
\usepackage{colortbl}
\usepackage{wrapfig}

\usepackage{fontawesome5}
\def\paperID{*****} 
\def\confName{CVPR}
\def\confYear{2026}
\makeatletter
\def\thanks#1{\protected@xdef\@thanks{\@thanks
        \protect\footnotetext{#1}}}
\makeatother
\title{IR-Flow: Bridging Discriminative and Generative \\ Image Restoration via Rectified Flow}
\author{Zihao Fan,~~Xin Lu,~~Jie Xiao,~~Dong Li,~~Jie Huang,~~Xueyang Fu$^\dagger$\\
MoE Key Laboratory of Brain-inspired Intelligent Perception and Cognition,\\
School of Information Science and Technology, University of Science and Technology of China \\
{~~ \tt\small fanzh03@mail.ustc.edu.cn, xyfu@ustc.edu.cn
}
}
\thanks{\hspace{-2mm} $\dagger$ : Corresponding author.  }

\begin{document}
\maketitle
\begin{abstract}
In image restoration, single-step discriminative mappings often lack fine details via expectation learning, whereas generative paradigms suffer from inefficient multi-step sampling and noise-residual coupling.
To address this dilemma, we propose IR-Flow, a novel image restoration method based on Rectified Flow that serves as a unified framework bridging the gap between discriminative and generative paradigms.
Specifically, we first construct multilevel data distribution flows, which expand the ability of models to learn from and adapt to various levels of degradation. 
Subsequently, cumulative velocity fields are proposed to learn transport trajectories across varying degradation levels, guiding intermediate states toward the clean target, while a multi-step consistency constraint is presented to enforce trajectory coherence and boost few-step restoration performance.
We show that directly establishing a linear transport flow between degraded and clean image domains not only enables fast inference but also improves adaptability to out-of-distribution degradations. 
Extensive evaluations on deraining, denoising and raindrop removal tasks demonstrate that IR-Flow achieves competitive quantitative results with only a few sampling steps, offering an efficient and flexible framework that maintains an excellent distortion-perception balance.
Our code is available at \url{https://github.com/fanzh03/IR-Flow}

\end{abstract}    
\begin{figure}[!t]
  \centering
  \includegraphics[width=1.0\columnwidth]{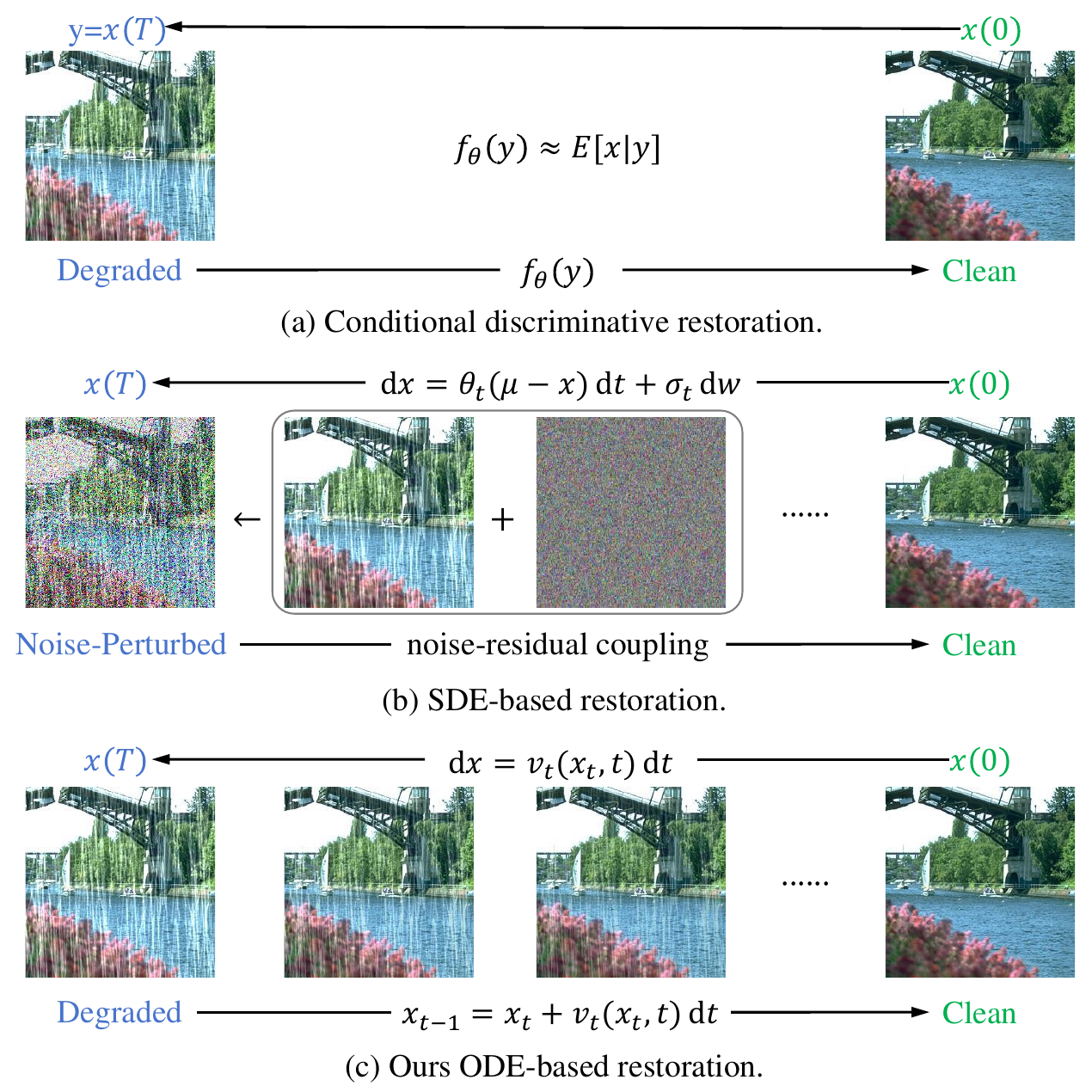}
  \vspace*{-1em}
  \caption{ Comparison of discriminative, SDE-based and our IR-Flow paradigms that bridges generative and discriminative image restoration. }
  \label{fig:contrast}
  \vspace*{-1em}
\end{figure}

\section{Introduction}
\label{sec:intro}
Image restoration involves recovering a clean image from its degraded counterpart~\cite{4MAXIM,SRCNN,9607618}.
In conventional discriminative learning for image restoration, given a dataset $\{(x_i, y_i)\}_{i=1}^N$ sampled from the joint distribution $p_{\text{data}}(x,y)$, the optimization objective is expressed as:
\begin{equation}
    \min_\theta \ \mathbb{E}_{(x,y)\sim p_{\text{data}}} \big[ \ell(f_\theta(y), x) \big],
\end{equation}
where $\ell(\cdot, \cdot)$ is a distortion-oriented loss, typically $\ell_2$ or $\ell_1$. Under the $\ell_2$ constraint, this formulation effectively constrains $f_\theta(x)$ to approximate the conditional expectation: $f_\theta(y) \approx \mathbb{E}[x\mid y]$, thus ensuring that the learned model minimizes pixel-wise reconstruction errors. While this leads to low distortion, it often produces overly smooth outputs, lacking perceptual realism.

\begin{figure*}[!t]
\setlength{\abovecaptionskip}{0.1cm}
  \begin{center}
    \includegraphics[width=0.8\linewidth]{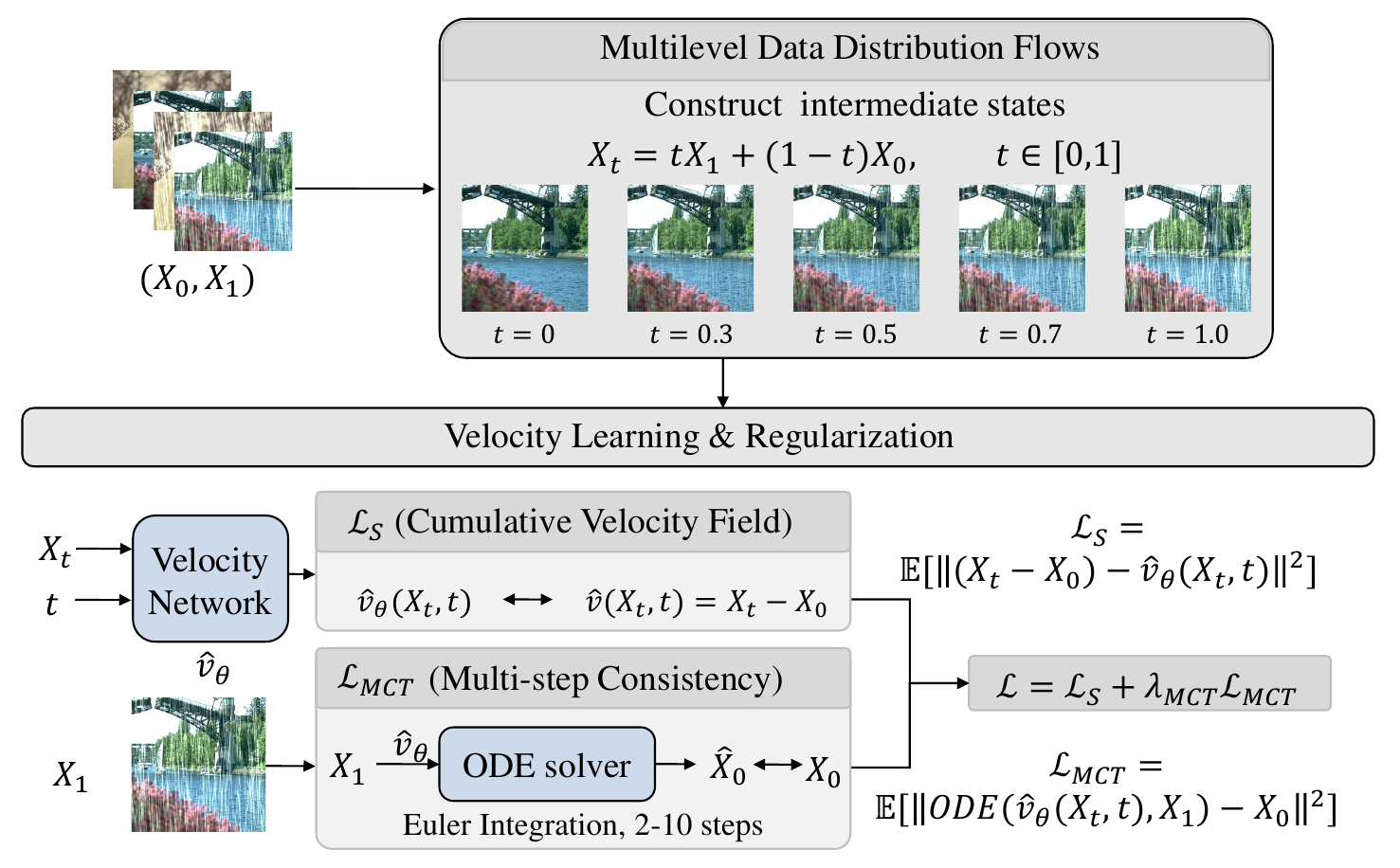}
  \end{center}
  \vspace*{-1em}
  \caption{Overall framework of IR-Flow. The approach achieves image restoration through Rectified Flow, refined by cumulative velocity field and multi-step consistency training.}
  \label{fig:pipeline}
  \vspace*{-1em}
\end{figure*}

In recent years, we have witnessed a paradigm shift from discriminative mappings to generative paradigms. Generative models (e.g., denoising diffusion probabilistic models~\cite{19DDPM,20DDIM}) serve as powerful tools for learning and sampling from complex data distributions, which focus on aligning the full data distribution by minimizing the Kullback-Leibler (KL) divergence:
\begin{equation}
    \min_\theta \ D_{\mathrm{KL}}\big(p_{\text{data}} \parallel p_\theta \big),
\end{equation}
where $p_\theta$ denotes the model distribution and $p_{\text{data}}$ the true data distribution. Unlike discriminative regression which enforces pixel-wise fidelity, this probabilistic formulation enables the model to capture high-frequency details, textures, and perceptual cues. 
Recent studies~\cite{ResShift,Resfusion,RDDM} have shown that formulating image restoration as a diffusion process, coupled with multi-step stochastic reverse denoising, yields highly effective results.
A more general approach is the recently proposed IR-SDE~\cite{IRSDE}, which models the image degradation process using mean-reverting stochastic differential equations (SDE) in Figure~\ref{fig:contrast}(b). Despite its effectiveness, it relies on carefully designed noise perturbation schedules that couple noise and residual learning (i.e., degraded inputs $y=x_1$, the target images $x_0$, and noise $\epsilon$):
\begin{equation}
    x_t=\alpha_tx_0+\beta_tx_1+\gamma_t\epsilon,
    \label{eq:sde}
\end{equation}
where $\alpha_t$, $\beta_t$ and $\gamma_t$ are control intensity, consequently increasing the number of sampling steps required for inference, highlighting the lack of a unified framework that harmonizes efficiency with the distortion-perception balance. 

To resolve this, we propose IR-Flow, a unified framework that bridges generative modeling and discriminative image restoration.
Motivated by the inefficiency of noise-driven diffusion formulations, we show that directly constructing a transport flow between degraded and clean images provides a more effective and physically interpretable solution for image restoration in Figure~\ref{fig:contrast}(c).
By learning a cumulative velocity field that captures the directional flow from degraded inputs to their clean counterparts, our ODE-based formulation enables more efficient inference while achieving performance comparable to SDE-based methods that jointly model noise and residual learning.
Moreover, the proposed cumulative velocity fields accelerate the convergence of velocity fields learning in standard rectified flow. 
Ultimately, by introducing a trajectory consistency constraint, we encourage the transport paths to align linearly, achieving a better balance between distortion and perceptual quality.

Our contributions are summarized as follows:
\begin{itemize}
\item 
We propose IR-Flow, a unified framework that bridges generative modeling and discriminative restoration by constructing a direct transport flow between degraded and clean images, which expands the model’s capability to handle diverse degradation levels.
\item 
We propose a cumulative velocity field that captures the directional transport from variably degraded states to their clean counterparts, while also accelerating the convergence of velocity learning for image generation.
\item 
We impose a trajectory consistency constraint to regularize intermediate states along straight paths, achieving a favorable balance between distortion fidelity and perceptual quality with only a few sampling steps.
\item 
Through extensive experiments on various representative image restoration tasks, our method demonstrates superior performance, while exhibiting strong generalization capabilities with varying degradation levels.
\end{itemize}

\begin{figure*}[!t]
\setlength{\abovecaptionskip}{0.1cm}
  \begin{center}
    \includegraphics[width=1.0\linewidth]{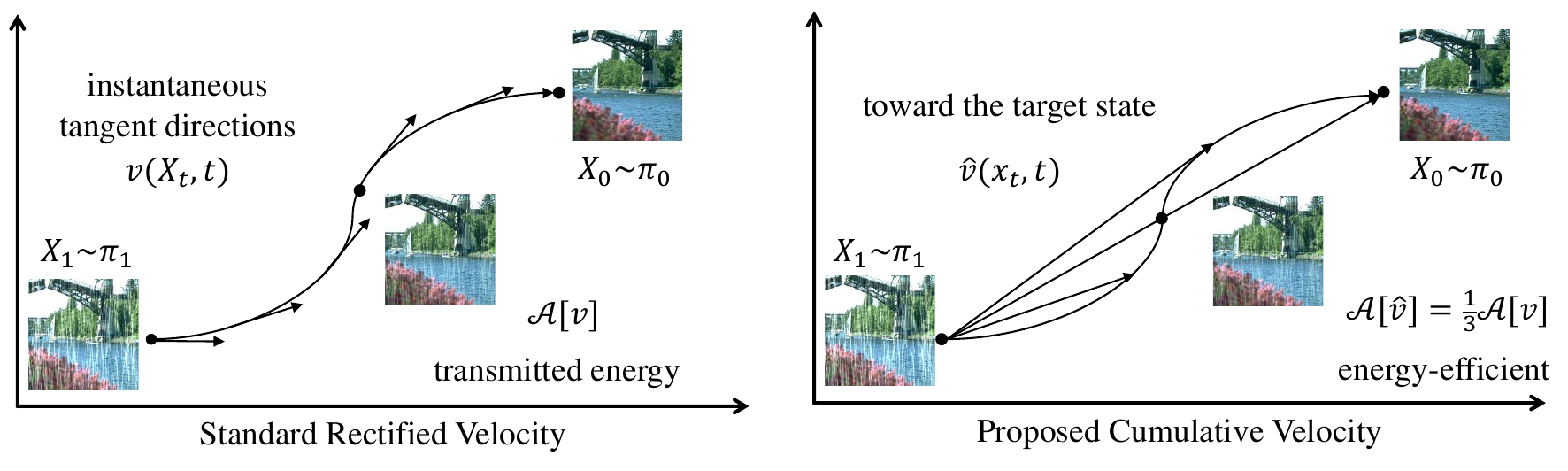}
  \end{center}
  \vspace*{-1em}
  \caption{Comparison of two velocity fields. Standard velocity fields represents the {instantaneous} tangent directions. Our representation captures {accumulated} velocity toward the target state.}
  \label{fig:velocity}
  \vspace*{-1em}
\end{figure*}

\section{Related Work}
\label{sec:related}
Traditional discriminative modeling approaches have achieved remarkable progress in classical restoration tasks, including image deraining~\cite{Derain1PReNet,derain2MPRNet,Derain3JORDER}, image denoising~\cite{denois7DnCNN,denois8FFDNet,Zhang2023}, raindrop removal~\cite{IDT,9009448,TransWeather} and image deblurring~\cite{Gopro,deblur14-DeblurGANv2,NAFNet}, as well as various other image restoration challenges~\cite{MIR-2022-12-398,Wu2024,MIR-2022-10-308,MIR-2024-03-083}.
In Transformer-based approaches, spatial self-attention is primarily used to capture long-range dependencies between pixels, such as SwinIR~\cite{SwinIR}, Uformer~\cite{16Uformer}, IPT~\cite{IPT} Restormer~\cite{5Restormer}, X-Restormer~\cite{XRestormer}, have achieved competitive performance across various degradation types. These methods all entail training a neural network to directly predict high-quality images from given low-quality ones. 

Generative diffusion models~\cite{19DDPM,20DDIM}, governed by SDEs~\cite{23scoreSDE,fang2024vividvideovirtualtryon,Lu2025,SD}, have achieved remarkable success, by leveraging a Markov chain to progressively corrupt an image into white Gaussian noise via a forward diffusion process, followed by the use of a deep neural network to approximate the reverse process for image reconstruction. 
Building upon this, Flow Matching~\cite{flowMatching,RectifiedFlow,SD3} extends the paradigm by modeling velocity fields to define transport paths between distributions. Attributed to its powerful generative capability, the diffusion model has shown considerable potential in addressing image restoration tasks.

Broadly speaking, existing approaches typically follow two main paradigms.
On the one hand, some methods~\cite{9887996,DiffIR,SRDiff,9879619} directly incorporate low-quality images as an addition condition into vanilla diffusion models~\cite{19DDPM} and retrain them specifically for the restoration task, or leverage pre-trained ones~\cite{SD,NEURIPS2021_49ad23d1,SD3} by employing degradation information to guide the reverse denoising process~\cite{DDPG,DiffPIR,DDRM,DDNM,DiffusionPrior2023,Dreamclean}.
On the other hand, alternative approaches(e.g., InDI~\cite{InDI}, Resfusion~\cite{Resfusion}, IR-SDE~\cite{IRSDE}, RDDM~\cite{RDDM}, and others~\cite{ResShift,I2SB,GOUB,zhu2025unidb}), reformulate the diffusion process and design new sampling trajectories by mixing three components: the degraded input $y$, the target image $x_0$, and noise $\epsilon$.
However, these methods either require multi-step SDE-based sampling starting from pure noise, or couple residual and noise learning, which exacerbates the optimization difficulty and ultimately leads to inefficient inference.

In recent work, several methods have focused on Flow-based modeling. While methods like FlowIE~\cite{FlowIE} and PMRF~\cite{PMRF} utilize rectified flows via diffusion-priors or two-stage posterior estimation, they lack a unified formulation.  In comparison to our approach, which directly establishes a flow between degraded and clean images and learns cumulative velocity fields, the two-stage methods lack a unified modeling framework.

\section{Method}
In this section, we present \textbf{IR-Flow} in Figure~\ref{fig:pipeline}, the key idea of our proposed image restoration approach is the transport mapping based on Rectified Flow.
Specifically, we formulate physically grounded and mathematically interpretable multilevel data distribution flows in Section~\ref{subsec:LDDF}. 
Building upon this foundation, we refine velocity field constraints through cumulative velocity field in Section~\ref{subsec:CVF}. 
Then we carefully construct the multi-step consistency training algorithm and ODE sampling paradigms in Section~\ref{subsec:train_sampling}.

\subsection{Multilevel Data Distribution Flows}
\label{subsec:LDDF}

The Rectified Flow~\cite{RectifiedFlow} addresses this by constructing straight paths between $\pi_0$ and $\pi_1$:
\begin{equation}
X_t=tX_1+(1-t)X_0,\ \ \ \ \ t\in [0,1], 
\end{equation}
where $X_t$ is the interpolated value at time $t$ and randomly paired $X_1\sim \mathcal{N}(\mathbf{0},\mathbf{I})$, $X_0\sim p_{data}$. The linear path is theoretically optimal as the shortest trajectory between endpoints. By learning the velocity field governing the transformation between the two distributions:
\begin{equation}
\frac{dX_t}{dt} = v(X_t,t),
\end{equation}
where $v(X_t,t)=\mathbb{E}[X_1-X_0\mid X_t]:\mathcal{X} \times [0,1] \rightarrow \mathbb{R}^d$ represents the velocity field guiding the flow to follow the direction of $(X_1-X_0)$, high-quality results can be achieved with a few Euler discretization step. 
The image restoration problem can thus be formulated as establishing a transport map between the degraded and clean image distributions. 
By solving a least-squares regression problem, the drift term is optimized to align with the direction of linear interpolation:
\begin{equation}
    \min_v \int_0^1 \mathbb{E} \big[ \| (X_1 - X_0) - v(X_t, t) \|^2 \big] dt,
\end{equation}
where $X_1\sim \pi_1$ represents degraded observations and $X_0\sim \pi_0$ their clean counterparts, with the loss function:
\begin{equation}
    \mathcal{L}=\mathbb{E}_{\ t,(X_0,X_1)}[\|(X_1-X_0)-v_\theta(X_t,t)\|^2],
\end{equation}
where $v_\theta$ represents parameterized velocity network of the velocity field $v$. 

Based on this framework, the multilevel data distribution flows $\pi_t$ can be constructed. 
Physically, for specific types of degradation (e.g., noise intensity, blur severity, or rain streak density), different images exhibit varying degrees of degradation. The creation of multilevel distribution flows artificially introduces intermediate states with controlled degradation levels, guiding the model to learn broader degradation scenarios. 
Mathematically, regions of low data density often lead to coarse distribution estimates due to sparse samples. By augmenting sparse regions with artificial interpolations, the local structure of the distribution is reinforced, enabling the model to better learn the velocity field for transitions between distributions. 
Additionally, fixing $t=1$ yields $X_{t=1} =X_1=y$, then $f_\theta(y)=v_\theta(X_1,1)$, which corresponds to discriminative paradigm and can be viewed as a special case of our approach. 
\subsection{Cumulative Velocity Field}
\label{subsec:CVF}
The standard Rectified Flow approach employs a velocity field defined as $v(X_t, t) = \mathbb{E}[X_1-X_0\mid X_t]$ to solve ordinary differential equations through multi-step numerical solvers. 
However, this approach of directly learning the transformation between distributions $\pi_0$ and $\pi_1$ creates an excessively large learning span, which results in three critical limitations: increased difficulty in estimating the velocity field, slower convergence during optimization, and suboptimal model fitting. 
To address the above limitations, we propose the cumulative velocity field (CVF) learning paradigm:
\begin{equation}
    \label{eq:Local-Velocity}
    \hat{v}(X_t,t)=\mathbb{E}[X_t-X_0 \mid X_t],
\end{equation}
\begin{equation}
    \label{eq:new-loss}
    \mathcal{L}_S=\mathbb{E}_{\ t,(X_0,X_1)}[\|(X_t-X_0)-v_\theta(X_t,t)\|^2],
\end{equation}
where the velocity field $\hat{v}(X_t,t)$ governs the intermediate states to clean distribution mapping.
Since the instantaneous velocity $v$ defines the tangent direction of the trajectory, a small step size is required to accurately integrate the path over multiple steps. By contrast, the cumulative velocity $\hat{v}$ points directly from an intermediate state toward the target, effectively mitigating directional deviation, as illustrated in Figure~\ref{fig:velocity}.

Compared to the standard constant velocity field used in rectified flow, the CVF is more energy-efficient in the sense of transport:
\begin{equation}
\mathcal{A}[\hat{v}] = \frac{1}{3} \, \mathcal{A}[v],
\end{equation}
where $\mathcal{A}$ represents the transmitted energy (more explanations are in the Appendix).
As a result, learning CVF significantly reduces the modeling complexity, accelerates convergence, and enables the model to achieve stronger empirical performance with substantially lower training cost. 
Empirical results (Section~\ref{sec:exp}) validate the advantages through comparative analysis of $\hat{v}(X_t,t)$ and $v(X_t,t)$. 
Notably, our formulation recovers the baseline velocity field as a special case when $t=1, i.e., \hat{v}(X_1 ,1) = v(X_t ,t)$, thereby establishing backward compatibility with standard rectified flow formulations. 

\begin{figure*}[!t]
\centering
\setlength{\abovecaptionskip}{0.1cm}
\includegraphics[width=0.8\textwidth]{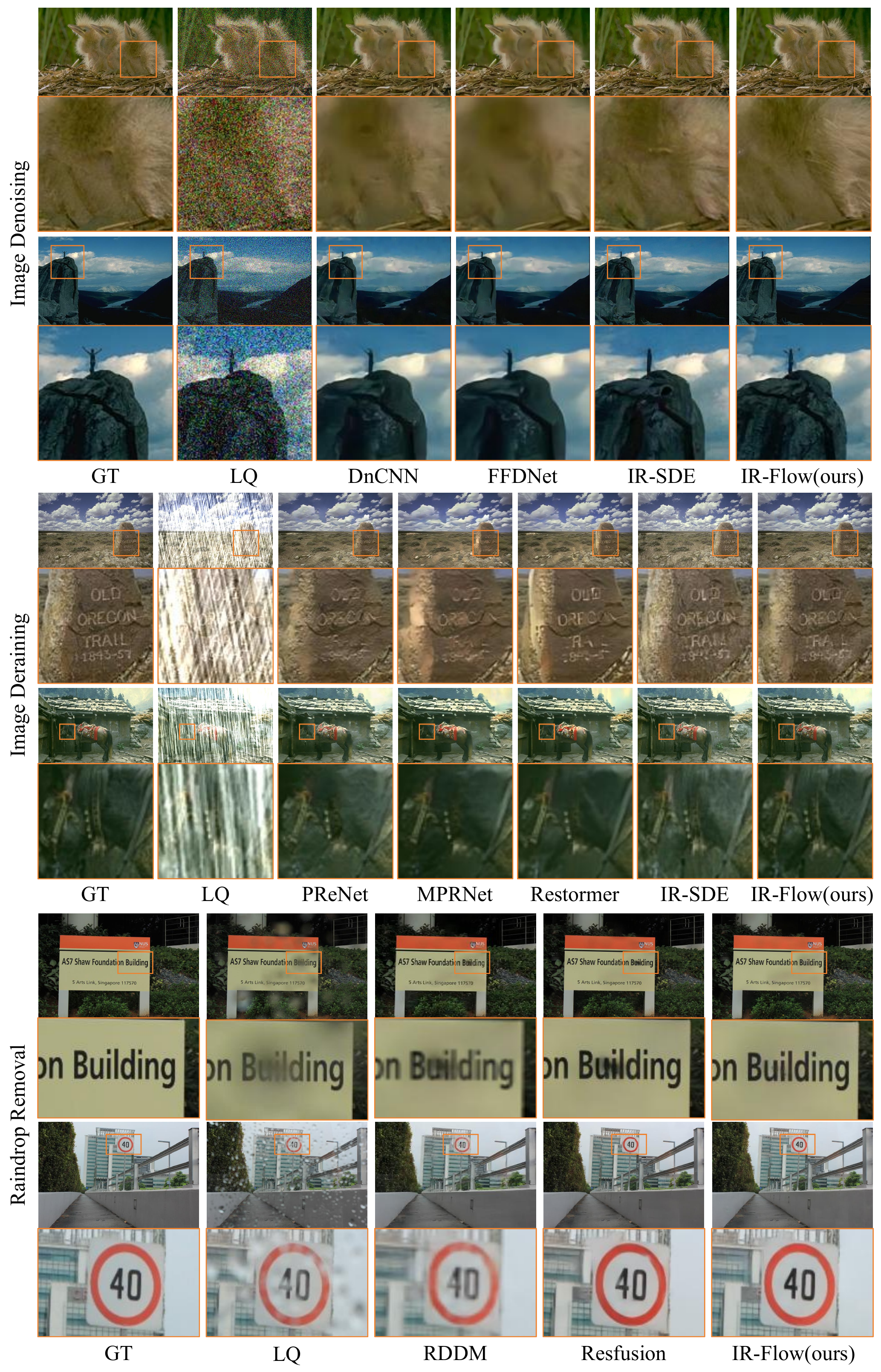}
\caption{Visual results of IR-Flow and competing approaches. The inference steps are set to 2 for deraining and raindrop removal, and 4 for denoising.}
\label{fig:main_pic}
\vspace*{-1em}
\end{figure*}

\subsection{Multi-step Consistency Training Loss}
\label{subsec:train_sampling}
To ensure that intermediate states during multi-step solving remain on the linear transport path between the source and target distributions, we introduce a multi-step consistency loss during training (MCT loss). Formally, this reconstruction-based regularization term is defined as:
\begin{multline}
    \mathcal{L}_{MCT} =
    \mathbb{E}_{(X_0,X_1)}\left[\|\mathrm{ODE}(\hat{v}_\theta(X_t,t),X_1) - X_0\|_2^2\right],
\end{multline}
where the ODE solver performs first-order Euler integration using 2 to 10 discretization steps to reconstruct the target. This explicitly constrains the velocity field direction to align with the linear path between distributions while maintaining intermediate states on the straight trajectory. The composite objective combines standard cumulative velocity matching loss $L_S$ with our consistency term:  
\begin{equation}
\mathcal{L} = \mathcal{L}_S + \lambda_{MCT} \mathcal{L}_{MCT},
\end{equation}
where the coefficient $\lambda_{MCT}=0.3$ is empirically selected to maintain commensurate scaling between the loss components.
For sampling, we design both first-order numerical solvers (Algorithm 1 and 2 in Appendix).

Notably, our analysis reveals an intrinsic perception-distortion tradeoff in multi-step sampling, increased steps with finer discretization enhance perceptual quality, while reduced steps with coarser intervals improve distortion metrics, consistent with theoretical predictions \cite{Tradeoff1,Tradeoff2}. Through systematic comparisons of order and step configurations, we propose a few-step sampling protocol that achieves optimal efficiency-quality equilibrium.

\begin{table*}[!htb]
\centering
\footnotesize
\setlength{\tabcolsep}{3pt}
\renewcommand{\arraystretch}{1.0}
\caption{Quantitative comparison on two deraining datasets.}
\vspace*{-1em}
 
\begin{tabularx}{\textwidth}{
@{}
>{\raggedright\arraybackslash}X
*{7}{>{\centering\arraybackslash}X}
@{}}
\toprule
\multirow{2}{*}{Method} 
& \multicolumn{3}{c}{Rain100H\cite{RainHL}} 
& \multicolumn{3}{c}{Rain100L\cite{RainHL}} 
& \multirow{2}{*}{NFE} \\
\cmidrule(lr){2-4} \cmidrule(lr){5-7}
& PSNR$\uparrow$ & SSIM$\uparrow$ & LPIPS$\downarrow$ 
& PSNR$\uparrow$ & SSIM$\uparrow$ & LPIPS$\downarrow$ & \\
\midrule
JORDER\cite{Derain3JORDER}      & 26.25 & 0.8349 & 0.197 & 36.61 & 0.9735 & 0.028 & 1 \\
PReNet\cite{Derain1PReNet}      & 29.46 & 0.8990 & 0.128 & 37.48 & 0.9792 & 0.020 & 1 \\
MPRNet\cite{derain2MPRNet}     & 30.41 & 0.8911 & 0.158 & 36.40 & 0.9653 & 0.077 & 1 \\
MAXIM\cite{4MAXIM}       & 30.81 & 0.9013 & 0.133 & 38.06 & 0.9770 & 0.048 & 1 \\
Restormer\cite{5Restormer}   & 31.46 & 0.9040 & 0.086 & 38.99 & 0.9780 & 0.013 & 1 \\
IR-SDE\cite{IRSDE}      & 31.65 & 0.9014 & \textbf{0.047} & 38.30 & 0.9805 & \underline{0.014} & 100 \\
Baseline    & 29.12 & 0.8824 & 0.153 & 33.17 & 0.9583 & 0.068 & 1 \\
\rowcolor{gray!20}
Ours        & \textbf{32.69} & \textbf{0.9257} & 0.073 & \underline{38.62} & \textbf{0.9823} & 0.014 & 1 \\
\rowcolor{gray!20}
Ours        & \underline{32.67} & \underline{0.9256} & \underline{0.065} & 38.40 & \underline{0.9818} & \textbf{0.013} & 2 \\
\bottomrule
\end{tabularx}
\label{tab:combined}
\vspace*{-1em}
\end{table*}

\begin{table*}[!htb]
\centering
\footnotesize
\setlength{\tabcolsep}{3pt} 
\caption{Quantitative comparison on three denoising datasets with $\sigma=25$.}
\vspace*{-1em}
\begin{tabularx}{\textwidth}{@{}>{\hsize=1.3\hsize}X
                             *{9}{>{\centering\arraybackslash\hsize=0.8\hsize}X}
                             >{\centering\arraybackslash\hsize=0.6\hsize}X@{}}
\toprule
\multirow{2}{*}{Method} 
& \multicolumn{3}{c}{McMaster\cite{McMaster}} 
& \multicolumn{3}{c}{Kodak24\cite{kodak}} 
& \multicolumn{3}{c}{CBSD68\cite{CBSD68}} 
& \multirow{2}{*}{NFE} \\
\cmidrule(lr){2-4} \cmidrule(lr){5-7} \cmidrule(lr){8-10}
& PSNR$\uparrow$ & SSIM$\uparrow$ & LPIPS$\downarrow$ 
& PSNR$\uparrow$ & SSIM$\uparrow$ & LPIPS$\downarrow$ 
& PSNR$\uparrow$ & SSIM$\uparrow$ & LPIPS$\downarrow$ 
& \\
\midrule
DnCNN\cite{denois7DnCNN}         
& 31.52 & 0.8692 & 0.101 
& 32.02 & 0.8763 & 0.129 
& 31.24 & 0.8830 & 0.109
& 1 \\
FFDNet\cite{denois8FFDNet}        
& 32.36 & 0.8861 & 0.103 
& 32.13 & 0.8779 & 0.140 
& 31.22 & 0.8821 & 0.121 
& 1 \\
IR-SDE\cite{IRSDE}        
& 32.39 & 0.8791 & \underline{0.055} 
& 32.14 & 0.8739 & 0.078 
& 31.14 & 0.8777 & 0.074 
& 22 \\
Baseline  
& 31.79 & 0.8697 & 0.122 
& 32.73 & 0.8666 & 0.161 
& 30.74 & 0.8661 & 0.162 
& 1 \\
\rowcolor{gray!20}
Ours 
& {\textbf{33.15}} & \underline{0.9031} & 0.075 
& \textbf{32.90} & 0.8919 & 0.107 
& \textbf{31.70} & \underline{0.8927} & 0.096 
& 1 \\
\rowcolor{gray!20}
Ours 
& \underline{33.11} & \textbf{0.9033} & 0.062 
& \underline{32.84} & \textbf{0.8927} & 0.088 
& \underline{31.63} & \textbf{0.8938} & 0.078 
& 2 \\
\rowcolor{gray!20}
Ours 
& 33.06 & 0.9021 & 0.057 
& 32.74 & \underline{0.8923} & \underline{0.084} 
& 31.60 & 0.8930 & \underline{0.071} 
& 3 \\
\rowcolor{gray!20}
Ours 
& 32.93 & 0.9001 & \textbf{0.049} 
& 32.65 & 0.8909 & \textbf{0.078} 
& 31.44 & 0.8922 & \textbf{0.068} 
& 4 \\
\bottomrule
\end{tabularx}
\label{tab:noise-results}
\vspace*{-1em}
\end{table*}

\begin{table}[htbp]
\centering
\footnotesize
\setlength{\tabcolsep}{3pt}
\renewcommand{\arraystretch}{1.0}
\caption{Quantitative comparison on the RainDrop\cite{RainDrop} dataset.}
\begin{tabularx}{\columnwidth}{@{} l *{4}{>{\centering\arraybackslash}X} @{}}
    \toprule
    Method & PSNR$\uparrow$ & SSIM$\uparrow$ & LPIPS$\downarrow$ & NFE \\
    \midrule
    IDT\cite{IDT} & 31.87 & 0.931 & \underline{0.057} & 1 \\
    WeatherDiff\cite{ozdenizci2023} & 32.43 & 0.933 & -- & 50 \\
    RDDM\cite{RDDM} & 32.51 & \textbf{0.956} & -- & 5 \\
    Resfusion\cite{Resfusion} & 32.66 & 0.938 & 0.060 & 5 \\
    \rowcolor{gray!20}
    Ours & \underline{32.75} & 0.939 & 0.066 & 1 \\
    \rowcolor{gray!20}
    Ours & \textbf{32.82} & \underline{0.941} & \textbf{0.056} & 2 \\
    \bottomrule
\end{tabularx}
\label{tab:RainD}
\vspace*{-1em}
\end{table}

\section{Experiments}
\label{sec:exp}

\subsection{Main Results}
In this section, we will validate the effectiveness of our proposed method on three tasks: image deraining, image denoising, raindrop removal.
Specifically, we will compare our method with some of the state-of-the-art methods on PSNR, SSIM~\cite{SSIM}, LPIPS~\cite{LPIPS} metrics and NFE. The NFE (Number of Function Evaluations) refers to the number of function evaluations required to generate an image or data. 
In the tables, the best and second-best quality scores of the evaluated methods are \textbf{highlighted} and \underline{underlined}.

\subsubsection{Comparative Experiment}
\label{subsubsec:main}

We comprehensively evaluate IR-Flow on three image restoration tasks: image deraining, image denoising, and raindrop removal. 
For image deraining, we use the synthetic Rain100H~\cite{RainHL} and Rain100L~\cite{RainHL} datasets. 
For image denoising, we train our models on mixed images from DIV2K~\cite{DIV2K}, Flickr2K~\cite{Flickr2K}, BSD500~\cite{BSD500}, and Waterloo Exploration~\cite{Waterloo}, and evaluate them on the McMaster~\cite{McMaster}, Kodak24~\cite{kodak}, and CBSD68~\cite{CBSD68} datasets. 
To further validate our method in real-world rainy scenarios, we conduct experiments on the Raindrop~\cite{RainDrop} dataset.
We compare our method with diverse state-of-the-art approaches across these tasks. These include supervised restoration models (PReNet~\cite{Derain1PReNet}, MPRNet~\cite{derain2MPRNet}, Restormer~\cite{5Restormer}, DnCNN~\cite{denois7DnCNN}, FFDNet~\cite{denois8FFDNet}, IDT~\cite{IDT}) and SDE/diffusion-based approaches (IR-SDE~\cite{IRSDE}, WeatherDiff~\cite{ozdenizci2023}, RDDM~\cite{RDDM}, Resfusion~\cite{Resfusion}).

The quantitative results are shown in Tables~\ref{tab:combined}, \ref{tab:noise-results}, and \ref{tab:RainD}, alongside the qualitative comparisons in Figure~\ref{fig:main_pic}. IR-Flow achieves highly competitive performance across both distortion and perception metrics. 
Our method yields SOTA (state-of-the-art) distortion metrics across all datasets, while attaining optimal perception performance within just 4 sampling steps.
Note that achieving absolute SOTA performance on a specific single task is not the main focus of this paper. Compared to other SDE-based and supervised methods, we focus more on proposing a unified framework that bridges supervised and generative restoration. Specifically, our method demonstrates methodological superiority by striking a balanced trade-off between computational efficiency, perception, and distortion under limited sampling steps.

\begin{figure}[!tbp]
\centering

\includegraphics[width=\columnwidth]{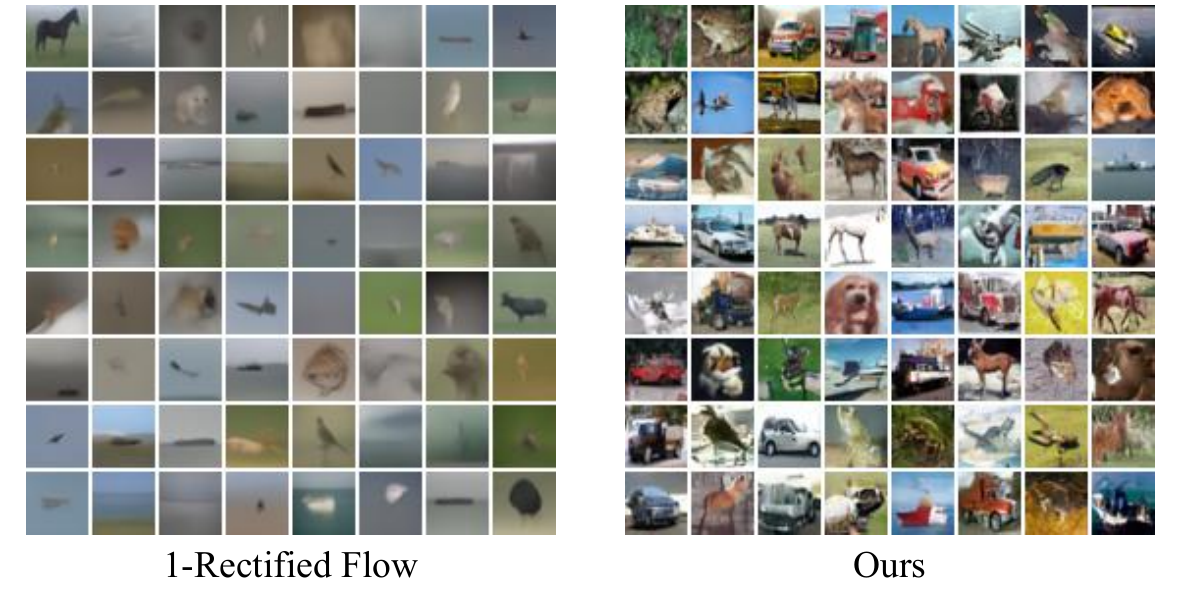}
\vspace*{-1em}
\caption{Visual comparisons between standard velocity and our velocity on CIFAR-10~\cite{cifar10} dataset with 50 steps.}
\label{fig:cifar}
\end{figure}

\subsection{Efficient Velocity for Image Generation}
We train DDIM~\cite{20DDIM}, standard Rectified Flow~\cite{RectifiedFlow}, and IR-Flow on the CIFAR-10~\cite{cifar10} dataset with the same backbone. We employ the FID~\cite{FID} as the quantitative metric. As shown in Table~\ref{table:cifar-results} and Figure~\ref{fig:cifar}, IR-Flow significantly outperforms standard Rectified Flow with the same sampling steps. Typically, rectified flow requires second-order rectification to achieve satisfactory results, our proposed cumulative velocity fields attain performance comparable to DDIM~\cite{20DDIM} without any rectification.

\subsection{Robustness and Generalization}
\begin{figure*}[!t]
\centering
\setlength{\abovecaptionskip}{0.1cm}
\includegraphics[width=1.0\textwidth]{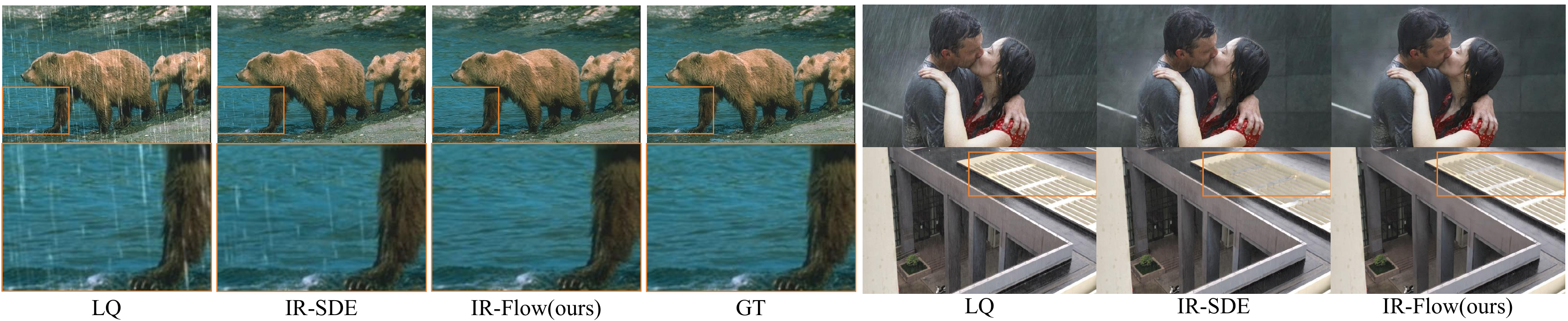}
\caption{Visual results of the model trained on Rain100H evaluated on the Rain100L and real-world Internet dataset for out-of-distribution (OOD) analysis. Our approach achieves higher quantitative metrics while producing cleaner rain-removal results.}
\label{fig:ood}
\end{figure*}

To evaluate the robustness and generalization of our model, we conduct experiments on denoising and deraining datasets. First, for the Gaussian noise experiment, we compare the unified training on $[0,50]$ with separate training on 15, 25, and 50, as shown in Table~\ref{table:app-12noise-results}. The results show that our method can adaptively learn to handle different noise levels; then for the deraining task, to validate the generalization of our method across different degradation levels, we train on Rain100H and evaluate on OOD datasets, including Rain100L and real-world Internet images, as shown in Table~\ref{tab:ood-combined} and Figure~\ref{fig:ood}. Since the real-world Internet dataset~\cite{luo2024controlling} lacks clean ground-truth images, we adopt no-reference metrics, including MUSIQ~\cite{MUSIQ}, NIMA~\cite{NIMA}, and LIQE~\cite{LIQE}. Compared with IR-SDE, which requires 100 NFEs, our method achieves superior performance with only a single NFE, producing cleaner details on OOD images. This can be attributed to our physically interpretable multilevel distribution flows, which effectively handle varying degradation levels, as well as the efficient cumulative velocity field.

\begin{table}[tbp]
\centering
\footnotesize
\caption{Quantitative comparisons with DDIM and Rectified Flow standard velocity on CIFAR-10~\cite{cifar10} dataset.}
\vspace*{-1em}
\begin{tabularx}{\columnwidth}{@{}c *{3}{>{\centering\arraybackslash}X}@{}}
    \toprule
    CIFAR\cite{cifar10} & DDIM & ${v}(X_t,t)$ & $\hat{v}(X_t,t)$ \\
    \midrule
    10 steps & \textbf{22.17} & 83.99 & 28.91 \\
    20 steps & 20.63 & 117.74 & \textbf{18.79} \\
    50 steps & 19.35 & 129.88 & \textbf{15.49} \\
    100 steps & 18.25 & 131.18 & \textbf{17.45} \\
    \bottomrule
\end{tabularx}
\label{table:cifar-results}
\vspace*{-1em}
\end{table}

\begin{table}[tbp]
\centering
\footnotesize
\setlength{\tabcolsep}{3pt}
\renewcommand{\arraystretch}{1.0}
\caption{Quantitative comparison on the DIV2K\cite{DIV2K} SR dataset.}
\vspace*{-1em}
\begin{tabularx}{\columnwidth}{@{} l *{4}{>{\centering\arraybackslash}X} @{}}
    \toprule
    Method & PSNR$\uparrow$ & SSIM$\uparrow$ & LPIPS$\downarrow$ & NFE \\
    \midrule
    InDI\cite{InDI} & 26.45 & -- & \textbf{0.136} & 100 \\
    IR-SDE\cite{IRSDE} & 25.90 & 0.6570 & 0.231 & 100 \\
    GOUB\cite{GOUB} & 26.89 & \underline{0.7478} & 0.220 & 100 \\
    UniDB\cite{zhu2025unidb} & 25.46 & 0.6856 & 0.179 & 100 \\
    \rowcolor{gray!20}
    Ours & \underline{27.03} & \textbf{0.7532} & 0.312 & 1 \\
    \rowcolor{gray!20}
    Ours & \textbf{27.04} & \textbf{0.7532} & 0.311 & 2 \\
    \bottomrule
\end{tabularx}
\label{tab:app-DIV2K}
\vspace*{-1em}
\end{table}

\begin{table}[htbp]
\centering
\footnotesize
\setlength{\tabcolsep}{3pt}
\renewcommand{\arraystretch}{1.0}
\caption{Quantitative comparison on the SIDD~\cite{SIDD} denoising dataset.}
\vspace*{-1em}
\begin{tabularx}{\columnwidth}{@{} l *{4}{>{\centering\arraybackslash}X} @{}}
    \toprule
    Method & PSNR$\uparrow$ & SSIM$\uparrow$ & LPIPS$\downarrow$ & NFE \\
    \midrule
    Restormer\cite{5Restormer} & \textbf{40.02} & \textbf{0.960} & 0.1977 & 1 \\
    \rowcolor{gray!20}
    IR-Flow (ours) & \underline{39.47} & \underline{0.918} & \underline{0.1931} & 1 \\
    \rowcolor{gray!20}
    IR-Flow (ours) & 39.36 & 0.916 & \textbf{0.1611} & 2 \\
    \bottomrule
\end{tabularx}
\label{tab:app-sidd}
\end{table}

\begin{table}[tbp]
\centering
\footnotesize
\setlength{\tabcolsep}{3pt}
\renewcommand{\arraystretch}{1.0}
\caption{Quantitative comparison on the RESIDE-6k\cite{hazedata} dehazing dataset.}
\vspace*{-1em}

\begin{tabularx}{\columnwidth}{@{} l *{4}{>{\centering\arraybackslash}X} @{}}
    \toprule
    Method & PSNR$\uparrow$ & SSIM$\uparrow$ & LPIPS$\downarrow$ & NFE \\
    \midrule
    IR-SDE\cite{IRSDE} & 25.25 & 0.906 & 0.060 & 100 \\
    \rowcolor{gray!20}
    IR-Flow (ours) & \textbf{25.68} & \textbf{0.943} & \underline{0.035} & 1 \\
    \rowcolor{gray!20}
    IR-Flow (ours) & \underline{25.39} & \underline{0.942} & \textbf{0.034} & 2 \\
    \bottomrule
\end{tabularx}
\label{tab:app-dehazing}
\vspace*{-1em}
\end{table}

\begin{figure}[htbp]
\centering
\includegraphics[width=\columnwidth]{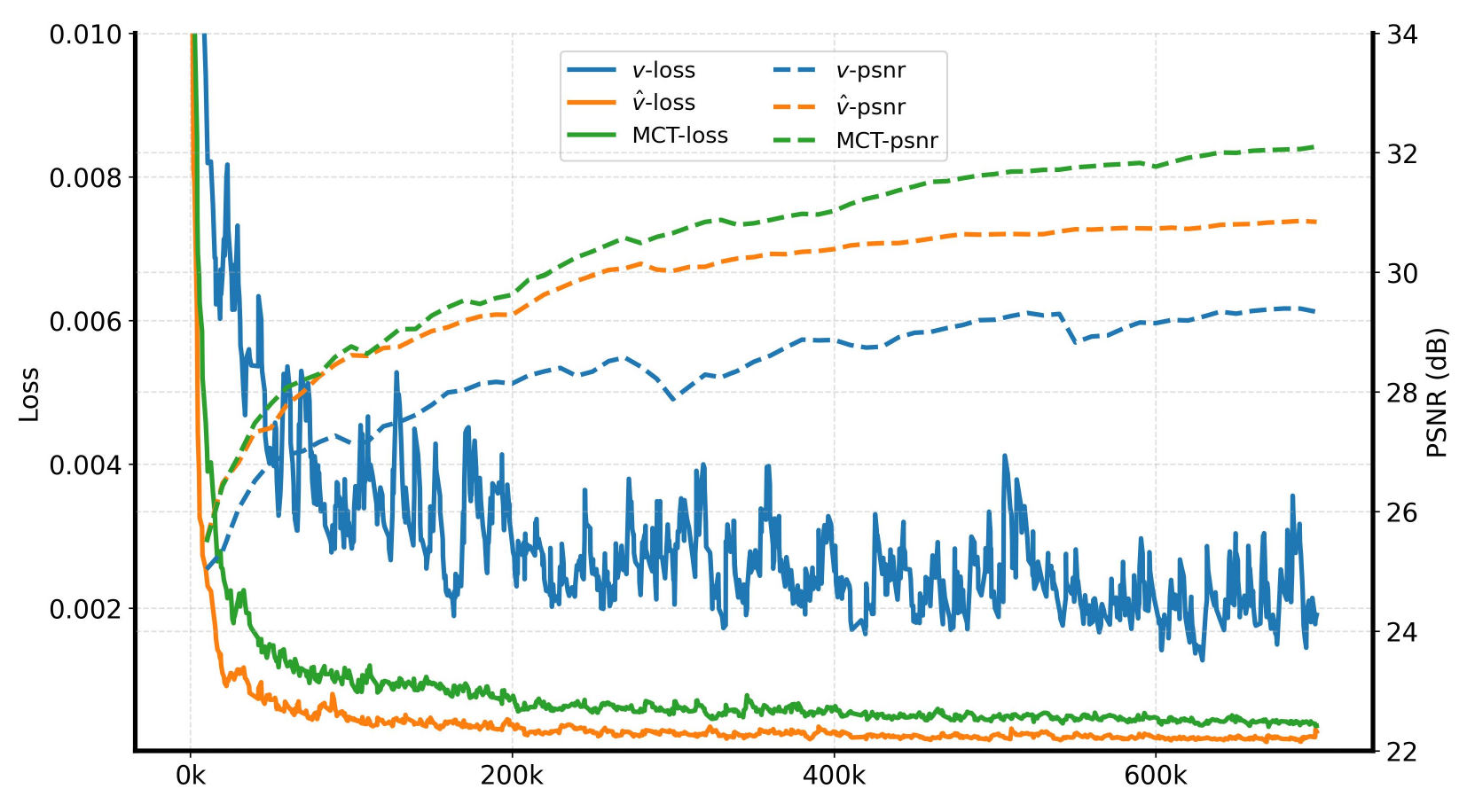}
\vspace*{-1em}
\caption{Training and Performance curves of the IR-Flow on progressive optimization methods.}
\label{fig:abl}
\end{figure}

\begin{table}[htbp]
\centering
\footnotesize
\setlength{\tabcolsep}{3pt}

\caption{Comparison of multi-step solution results.}
\vspace*{-1em}

\begin{tabularx}{\columnwidth}{@{}c *{3}{>{\centering\arraybackslash}X}@{}}
    \toprule
    NFE & ${v}(X_t,t)$ & $\hat{v}(X_t,t)$ & MCT \\
        & PSNR$\uparrow$/LPIPS$\downarrow$  &PSNR$\uparrow$/LPIPS$\downarrow$      & PSNR$\uparrow$/LPIPS$\downarrow$ \\
    \midrule
    1 & {\textbf{29.12}}/0.131 & \textbf{31.50}/\textbf{0.087} & {\textbf{32.69}}/0.073 \\
    2 & 29.10/0.122 & 30.85/0.094 & 32.67/0.065 \\
    3 & 28.97/0.119 & 30.67/0.094 & 32.57/0.063 \\
    4 & 28.87/0.118 & 29.70/0.102 & 32.53/0.060 \\
    5 & 28.74/\textbf{0.117} & 29.09/0.106 & 32.55/\textbf{0.059} \\
    \bottomrule
\end{tabularx}
\label{table:abl}
\end{table}

\begin{figure*}[!t]
\centering
\setlength{\abovecaptionskip}{0.1cm}
\includegraphics[width=0.85\textwidth]{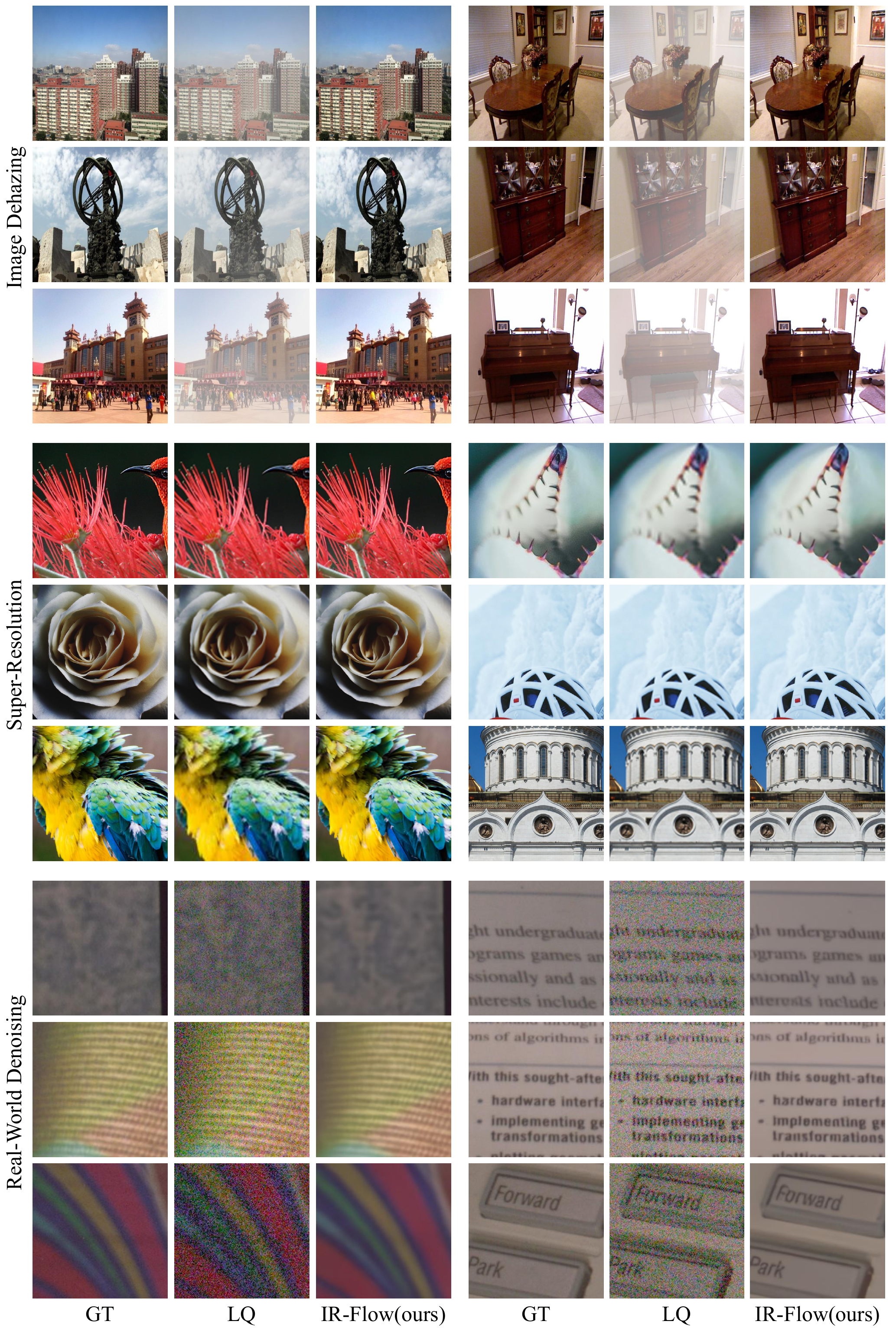}
\caption{Visual results of our IR-Flow method of dehazing on the RESIDE-6k~\cite{hazedata} dataset, 4$\times$ Super-Resolution on the DIV2K~\cite{DIV2K} dataset and real-world image denoising on SIDD~\cite{SIDD} dataset.}
\label{fig:supp_all}
\vspace*{-2em}
\end{figure*}

\begin{table*}[t]
\centering
\footnotesize
\caption{Quantitative comparison on gaussian denoising. ``Ours (separate)'' learns a separate model for each noise level $\sigma\in\{15,25,50\}$, while ``Ours (unified)'' learns a single model to handle various noise levels $\sigma\in[0,50]$.}
\vspace*{-1em}
\begin{tabularx}{\textwidth}{@{}l *{9}{>{\centering\arraybackslash}X} >{\centering\arraybackslash}X @{}}
\toprule
\multirow{2}{*}{Method} 
& \multicolumn{3}{c}{McMaster\cite{McMaster}} 
& \multicolumn{3}{c}{Kodak24\cite{kodak}} 
& \multicolumn{3}{c}{CBSD68\cite{CBSD68}} 
& \multirow{2}{*}{NFE} \\
\cmidrule(lr){2-4} \cmidrule(lr){5-7} \cmidrule(lr){8-10}
& $\sigma{=}15$ & $\sigma{=}25$ & $\sigma{=}50$ 
& $\sigma{=}15$ & $\sigma{=}25$ & $\sigma{=}50$ 
& $\sigma{=}15$ & $\sigma{=}25$ & $\sigma{=}50$ 
& \\
\midrule
Baseline 
& 33.51 & 31.79 & 29.15 
& 33.92 & 32.73 & 28.89 
& 33.02 & 30.74 & 27.84 
& 1 \\
\rowcolor{gray!20}
Ours (separate) 
& 35.40 & 33.15 & 30.13 
& 35.31 & 32.90 & 29.90 
& 34.31 & 31.70 & 28.52 
& 1 \\
\rowcolor{gray!20}
Ours (unified) 
& 35.38 & 33.11 & 30.02 
& 35.31 & 32.87 & 29.82 
& 34.30 & 31.68 & 28.46 
& 1 \\
\bottomrule
\end{tabularx}
\label{table:app-12noise-results}
\end{table*}

\begin{table*}[!htbp]
\centering
\footnotesize
\setlength{\tabcolsep}{3pt}
\caption{Out-of-distribution evaluation on Rain100L and Internet datasets.}
\vspace*{-1em}
\begin{tabularx}{\textwidth}{@{} l *{6}{>{\centering\arraybackslash}X} c @{}}
\toprule
\multirow{2}{*}{Method} & \multicolumn{3}{c}{Rain100L\cite{RainHL}} & \multicolumn{3}{c}{Real-World Internet~\cite{luo2024controlling}} & \multirow{2}{*}{NFE} \\
\cmidrule(lr){2-4} \cmidrule(lr){5-7}
& PSNR$\uparrow$ & SSIM$\uparrow$ & LPIPS$\downarrow$ & MUSIQ$\uparrow$ & NIMA$\uparrow$ & LIQE$\uparrow$ & \\
\midrule
IR-SDE\cite{IRSDE}         & 37.62 & 0.976 & \textbf{0.017} & 56.87 & 4.598 & 2.7112 & 100 \\
IR-Flow (Ours) & \textbf{38.17} & \textbf{0.981} & 0.019 & \textbf{57.14} & \textbf{4.669} & \textbf{2.7128} & 1 \\
\bottomrule
\end{tabularx}
\label{tab:ood-combined}
\end{table*}

\subsection{Evaluation on Various Tasks}
\label{subsec:app-exp}
In this section, we further validate the general applicability of the proposed IR-Flow by conducting qualitative experiments on image super-resolution, deblurring and dehazing. For single-image super-resolution, we train and evaluate our model on the DIV2K~\cite{DIV2K} dataset. For image dehazing, we adopt the RESIDE-6K~\cite{hazedata} dataset. The qualitative comparisons are presented in Table~\ref{tab:app-DIV2K}, Table~\ref{tab:app-dehazing} and Figure~\ref{fig:supp_all}, demonstrating the effectiveness of IR-Flow across different restoration tasks. We evaluate the real-world deblurring performance of IR-Flow on the GoPro\cite{Gopro} dataset. We also evaluate on real-world denoising datasets SIDD~\cite{SIDD}, comparing with Restormer\cite{5Restormer}, as shown in Figure~\ref{fig:supp_all} and Table~\ref{tab:app-sidd}.The related experimental results are discussed in Appendix.

\subsection{Ablation Experiments}
\label{subsec:ablation}

In addition to proposing a unified perspective for analyzing generative modeling and discriminative image restoration, the innovation of our article in modeling optimization lies in CVF and MCT loss. To verify the effectiveness of the method in learning optimization, we conducted ablation experiments using deraining as a case study. As shown in Figure~\ref{fig:abl} and Table~\ref{table:abl}, CVF $\hat{v}(X_t,t)$ accelerates model convergence and achieves better performance while ensuring training stability. After further introducing the MCT loss, it still maintains considerable stability and convergence speed, achieving even higher performance in multi-step solving. For MCT loss weight $\lambda_{MCT}$, we experimented with multiple values and empirically found that the best performance is achieved around $0.3$, as shown in Table~\ref{table:app-mct} in Appendix.

\subsection{Resource efficiency}
We compare the parameters, MACs (multiply-accumulate operations) and inference time with other image restoration methods, using 256 $\times$ 256 images as the input. As shown in Table~\ref{table:complexity_comparison}, IR-Flow uses the same network architecture as IR-SDE, it runs 4$\times$ faster than Resfusion~\cite{Resfusion}, 7$\times$ faster than RDDM~\cite{RDDM}, and even 100$\times$ faster than IR-SDE\cite{IRSDE} at inference time. This substantially accelerates the inference process, achieving a favorable balance between efficiency and performance. All experiments were conducted on an NVIDIA 2080Ti GPU.

\begin{table}[h]
\centering
\footnotesize
\caption{Resource efficiency and performance analysis.}
\vspace*{-1em}
\begin{tabular}{lcccc}
\toprule
Method & RDDM\cite{RDDM} & Resfusion\cite{Resfusion}  & IR-SDE\cite{IRSDE} & Ours \\
\midrule
Time(s) & 0.52 & 0.26 &  7.41 & 0.07 \\
Params(M)  & 15.5 & 7.7 &  137.1 & 137.1 \\
MACs(G) & 329.3  & 164.5& 378.7 & 378.7 \\
\bottomrule
\end{tabular}
\label{table:complexity_comparison}
\end{table}

\section{Conclusion}
\label{sec:Conclusion}

In this work, we propose IR-Flow, an efficient image restoration framework based on rectified flow that bridges discriminative and generative paradigms. By learning a linear transport flow between degraded and clean image domains, IR-Flow enables fast ODE-based inference and improves robustness to out-of-distribution degradations. We further present multilevel data distribution flows, cumulative velocity fields, and a multi-step consistency constraint to model straight trajectories and enhance few-step restoration. In addition, the cumulative velocity fields accelerates rectified flow training and demonstrates effectiveness in image generation task, highlighting the generality of our approach.

{
    \small
    \bibliographystyle{ieeenat_fullname}
    \bibliography{main}
}

\clearpage
\setcounter{page}{1}

\section{Implementation details}
\label{sec:app-exp_setting_dataset}
We chose the same U-Net backbone network as used in IR-SDE~\cite{IRSDE}. The CNN-baseline uses the same network but directly input the low-quality image and output the high-quality image. For most tasks, we set the training patch-size to be $256\times 256$ and use a batch size of $8$. We used the Adam~\cite{adam} optimizer with parameters $\beta_1 = 0.9$ and $\beta_2 = 0.99$. The total training steps are fixed to $700$ thousand and the initial learning rate set to $1e^{-4}$ and decays half per 200 thousand iterations. Our models are trained on single NVIDIA A40 GPU for each task. The training hyperparameters are set as $\lambda_{MCT}=0.3$. We set the time scheduling to be uniformly linear $t\in U(0,1)$, we set the default training timestep to $T=1000$, which corresponds to a discrete time interval of $0.001$.

\section{Unified Formulation of SDE-based Methods}
\label{sec:sde}
To unify the representation of various SDE-based generative formulations for restoration, 
we introduce a general parameterization of the noisy latent variable $x_t$ as:
\begin{equation}
    x_t = \alpha_t\, x_0 + \beta_t\, x_1 + \gamma_t\, \epsilon,
\end{equation}
where $x_0$ is the clean data, $x_1$ is the observed degraded image, $\epsilon \sim \mathcal{N}(0, I)$
is Gaussian noise, and $\alpha_t, \beta_t, \gamma_t$ are time-dependent coefficients. Including InDI~\cite{InDI}, I2SB~\cite{I2SB}, RDDM~\cite{RDDM}, ResShift~\cite{ResShift}, Resfusion~\cite{Resfusion} and IR-SDE~\cite{IRSDE}, can be recovered from this unified
parameterization by specifying different schedules of 
$(\alpha_t, \beta_t, \gamma_t)$, and their differences arise mainly from 
the choice of interpolation and noise injection strategies. Under the unified formulation, these SDE-based methods can be regarded as a form of \emph{noise–residual coupled learning}:
\begin{align}
    x_t &= \alpha_t\, x_0 + \beta_t\, x_1 + \gamma_t\, \epsilon, \\
    &=(\alpha_t\,+ \beta_t)\, x_0 \, +\,\beta_t\,(x_1-x_0)\, +\, \gamma_t\,\epsilon, \\
    &=(\alpha_t\,+ \beta_t)\, x_0 \, +\,\beta_t\,R\, +\, \gamma_t\,\epsilon.
\end{align}

Our method simplifies the noise–residual coupling and enables the learning of a velocity field with clear physical interpretation. We first construct multilevel data distribution flows, which expand the ability of models to learn from and adapt to various levels of degradation. Subsequently, the cumulative velocity fields are introduced to learn transport trajectories over varying degradation levels, guiding intermediate states toward the clean target.

\section{Comparing CVF and Standard Rectified Flow in Optimal Transport Energy}
\label{sec:ilvf-energy}

We provide a quantitative comparison between the \textbf{Cumulative Velocity Field} (CVF) and the standard Rectified Flow field in terms of transport energy under the Benamou--Brenier dynamic formulation of optimal transport.

\textbf{Setup.} 
Given paired samples $(X_0, X_1) \sim \pi_0 \times \pi_1$, define the linear interpolation:
\begin{equation}
X_t = (1 - t) X_0 + t X_1, \quad t \in [0, 1].
\end{equation}
A velocity field $v(X_t, t)$ drives the interpolated dynamics via $\frac{dX_t}{dt} = v(X_t, t)$, and the associated transport energy (action) is defined as:
\begin{equation}
\mathcal{A}[v] = \int_0^1 \mathbb{E} \left[ \|v(X_t, t)\|^2 \right] dt.
\end{equation}

\textbf{Case 1: Standard Rectified Flow.}
Using the standard velocity field $v(X_t, t) = X_1 - X_0$, which is constant across $t$:
\begin{equation}
\mathcal{A}[v] = \int_0^1 \mathbb{E} \left[ \|X_1 - X_0\|^2 \right] dt 
= \mathbb{E} \left[ \|X_1 - X_0\|^2 \right].
\end{equation}

\textbf{Case 2: Cumulative Velocity Field (CVF).}
Under CVF, the velocity field is defined as:
\begin{equation}
\hat{v}(X_t, t) = X_t - X_0 = t(X_1 - X_0).
\end{equation}
The corresponding energy becomes:
\begin{align*}
\mathcal{A}[\hat{v}] 
&= \int_0^1 \mathbb{E} \left[ \|t (X_1 - X_0)\|^2 \right] dt \\
&= \mathbb{E} \left[ \|X_1 - X_0\|^2 \right] \cdot \int_0^1 t^2 dt \\
&= \frac{1}{3} \, \mathbb{E} \left[ \|X_1 - X_0\|^2 \right].
\end{align*}

\textbf{Conclusion.}
The CVF formulation achieves a \textbf{3$\times$ reduction} in kinetic energy:
\begin{equation}
   \mathcal{A}[\hat{v}] = \frac{1}{3} \, \mathcal{A}[v]. 
\end{equation}

This reduction demonstrates that CVF is energetically more efficient, producing smoother trajectories and minimizing numerical discretization errors, which are crucial for achieving high-quality and fast image restoration.

\begin{table*}[t]
\centering
\footnotesize
\setlength{\tabcolsep}{4pt}
\caption{Quantitative comparison results of different $\lambda_{MCT}$.}
\begin{tabular*}{\textwidth}{@{\extracolsep{\fill}} c *{10}{c} @{}}
\toprule
Metrics & $0.1$ & $0.2$ & $0.3$ & $0.4$ & $0.5$ & $0.6$ & $0.7$ & $0.8$ & $0.9$ & $1.0$ \\
\midrule
PSNR & 28.59 & 28.43 & \underline{28.75} & 28.25 & 28.07 & 27.18 & 27.22 & 28.38 & 28.60 & \textbf{28.84} \\
SSIM & 0.8747 & 0.8743 & \textbf{0.8770} & \underline{0.8760} & 0.8694 & 0.8462 & 0.8581 & 0.8742 & 0.8753 & 0.8752 \\
LPIPS & 0.1157 & 0.1216 & \underline{0.1156} & 0.1247 & 0.1264 & 0.1373 & 0.1396 & 0.1183 & \textbf{0.1148} & 0.1202 \\
\bottomrule
\end{tabular*}
\label{table:app-mct}
\end{table*}

\section{Multi-step Consistency Training Algorithm}
\label{sec:app-Training-MCT}

Our method only requires modification of the paired dataset $D$, making it applicable to all supervised learning restoration tasks, $X_1\sim \pi_1$ represents degraded observations and $X_0\sim \pi_0$ their clean counterparts. For ODE solver, we choose a first-order numerical solver, as shown in Algorithm~\ref{alg:MCT} and \ref{alg:app-Euler-solver1}.

\section{Multi-step ODE Sampling Algorithms and Results}
\label{sec:app-Sampling-ODE}
For the CVF in this paper, we designed corresponding first-order solving algorithms, as shown in Algorithm~\ref{alg:app-Euler-solver1} below. Considering sampling efficiency, we use a first-order algorithm to solve all tasks.

\begin{figure*}[!htb]
  \begin{center}
    \includegraphics[width=1.0\linewidth]{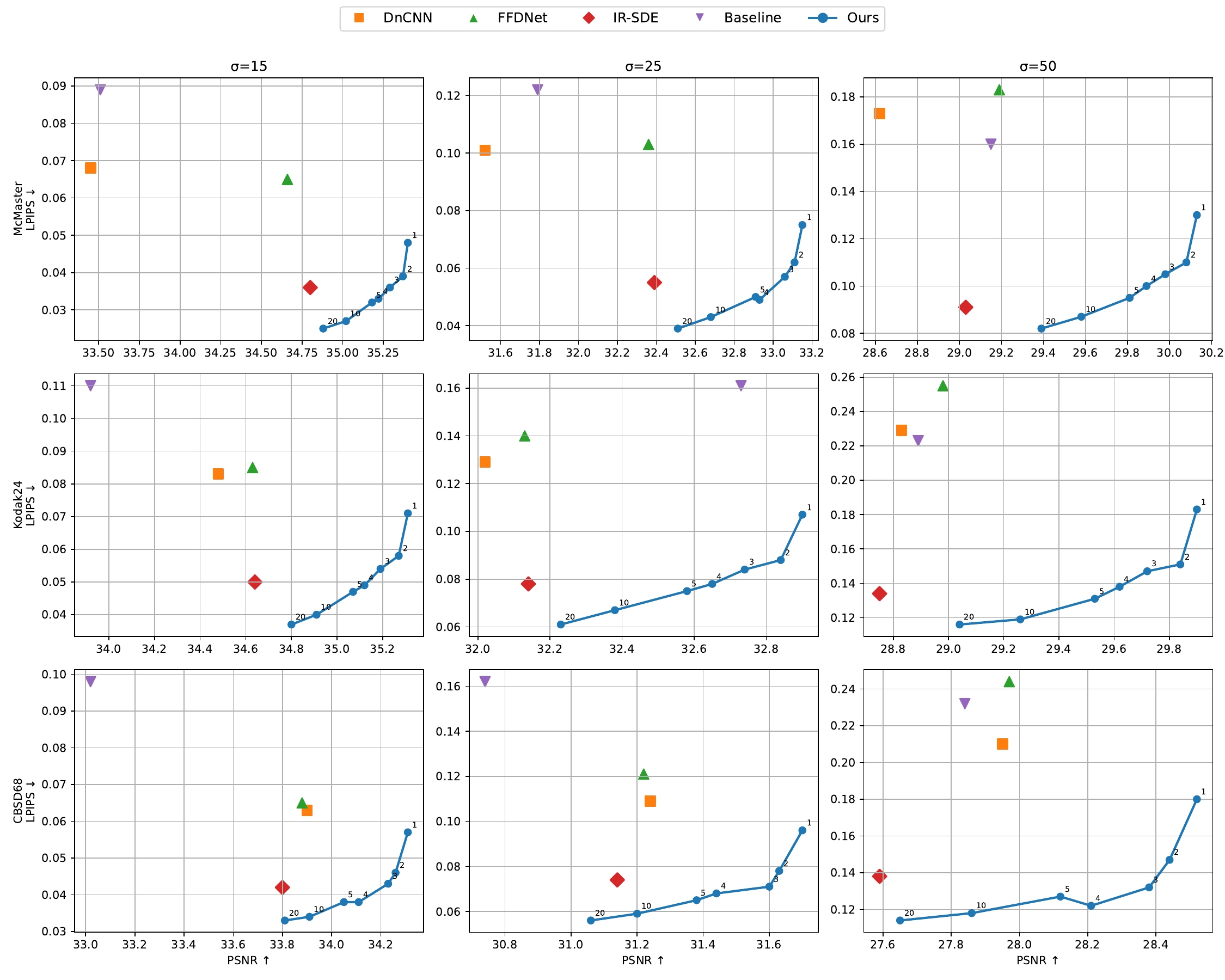}
  \end{center}
  \caption{Multi-step solution results on denoising. As the number of solving steps increases, the details and textures of the recovery results gradually become richer.}
  \label{fig:app-pareto}
\end{figure*}

\begin{figure*}[!htb]
  \begin{center}
    \includegraphics[width=1.0\linewidth]{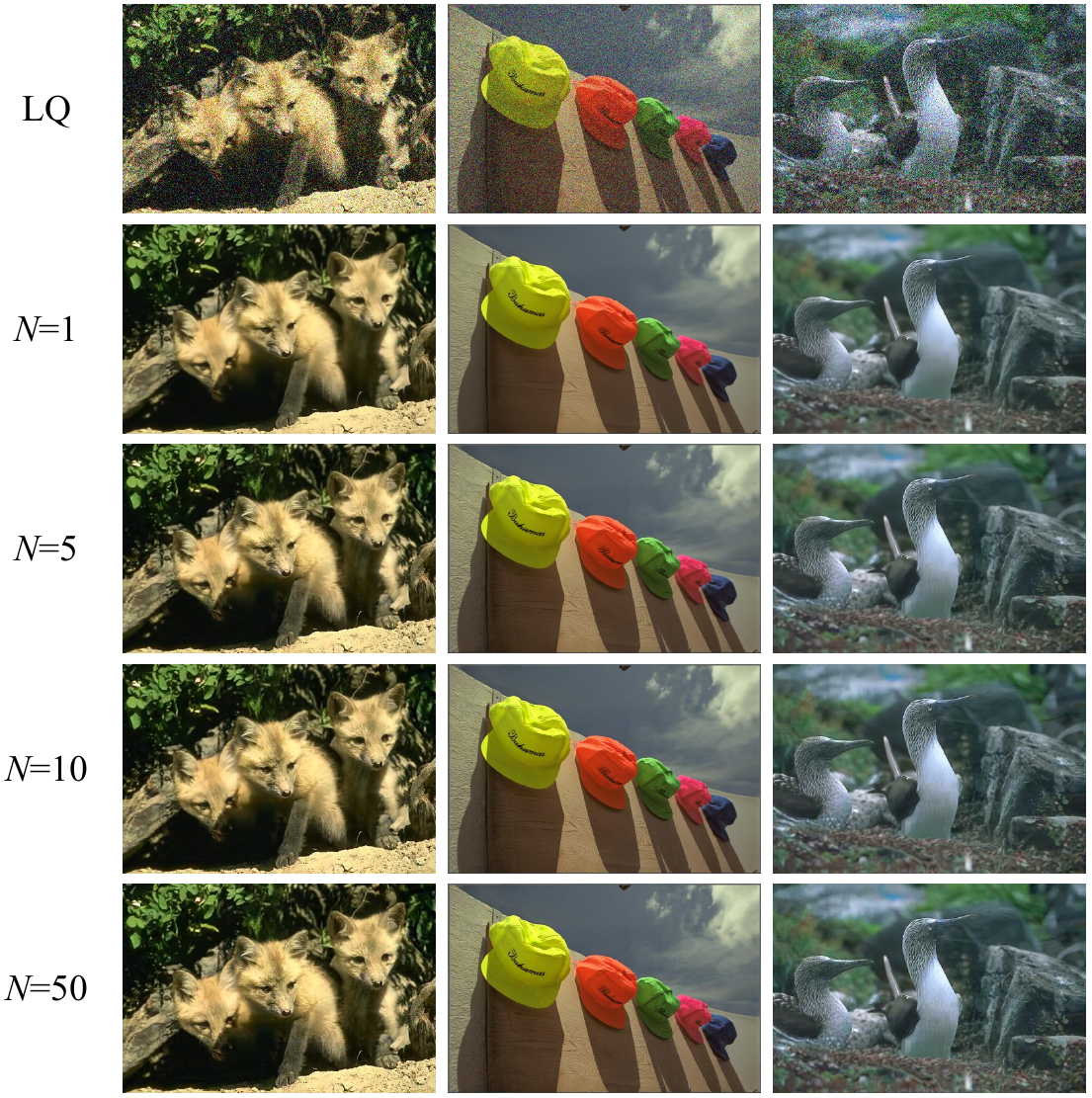}
  \end{center}
  \caption{Multi-step solution results on denoising. As the number of solving steps increases, the details and textures of the recovery results gradually become richer. (Zoom in visualization for detailed inspection of local structures. For example: fur, texture, etc.)}
  \label{fig:app-more-multi}
\end{figure*}

The comprehensive results of image denoising on three different test sets over different noise levels are given by Tables~\ref{table:app-multi-step-noise-level-M}, \ref{table:app-multi-step-noise-level-K}, and \ref{table:app-multi-step-noise-level-C}. 
The tabular data demonstrates that distortion metrics decline while perceptual quality improves with increasing optimization steps, leading our method to strike an optimal balance by constraining the step count to no more than 4 in the final output in Figure~\ref{fig:app-pareto}. As shown in the Figure~\ref{fig:app-more-multi}, details become progressively richer as the number of sampling steps increases.

\begin{algorithm}
\caption{Multi-step Consistency Training}
\label{alg:MCT}
\begin{algorithmic}[1]
\Require 
    Dataset $D$, model $v_\theta(x_t,t)$, MCT weight $\lambda_{\text{MCT}}$
\Ensure Optimized parameters $\theta$
\While{not converged}
    \State Sample $(x_{1}, x_{0}) \sim D$,\ \ $ t \sim \mathcal{U}[0, 1]$
    \State Construct Flows $x_{t} = t x_{1} + (1-t) x_{0}$
    \State Compute loss:
    $$
    \mathcal{L}_{S}(\theta) = \|\hat{v}_\theta(x_t,t)-(x_t-x_0)\|^2
    $$
    $$
    \mathcal{L}_{MCT}(\theta) = \|ODE\_solver(\hat{v}_\theta(x_t,t),x_1)-x_0\|^2
    $$
    $$
    \mathcal{L}(\theta) = \mathcal{L}_{S} + \lambda_{\text{MCT}}\mathcal{L}_{MCT}
    $$
\EndWhile
\end{algorithmic}
\end{algorithm}

\begin{algorithm}
\caption{ODE solver: $1^{st}$-order Euler}
\label{alg:app-Euler-solver1}
\begin{algorithmic}[1]
\Require 
    Dataset $D$, model $v_\theta(x_t,t)$, step $N$
\State Sample $x_{1} \sim D$
\State initialize $h=T/N$
\For{$i = N$ \textbf{to} $1$} 
    \State step  \ \ \ \ \ $j=i/N$
    \State weight  \ \ \ \ \ $w_i=1/i$
    \State velocity \ \ \ \ \ $d_i=\hat{v}_\theta(x_j,i\cdot h)$ 
    \State compute \ \ \ \ \ $x_{j-\frac{1}{N}}=x_j-w_i\cdot d_i$ 
\EndFor
\end{algorithmic}
\end{algorithm}

\begin{table}[!htbp]
\centering
\footnotesize
\caption{Quantitative comparison on the GoPro deblurring dataset.}
\vspace*{-1em}
\begin{tabular}{lccccc}
\toprule
Method & PSNR$\uparrow$ & SSIM$\uparrow$ & LPIPS$\downarrow$ & NFE \\ 
\midrule
  DeepDeblur    & 29.08 & 0.9135 & 0.135 & 1 \\
  DeblurGAN-v2  & 29.55 & \textbf{0.9340} & 0.117 &  1 \\
  DiffUIR       & 30.63 & 0.8900 & --  & 3 \\
  IR-SDE        & 30.70 & 0.9010 & \textbf{0.064} & 100 \\
  \rowcolor{gray!20}
  IR-Flow(ours)          & \textbf{31.20} & \underline{0.9149} & 0.115  & 1 \\
  \rowcolor{gray!20}
  IR-Flow(ours)          & \underline{31.15} & 0.9141 & \underline{0.109}  & 2 \\
\bottomrule
\end{tabular}
\label{table:GoProD}
\end{table}

\begin{table*}[!htb]
\begin{minipage}{\linewidth}
    \centering
        \caption{Quantitative results of image denoising on McMaster~\cite{McMaster} test set. Our method achieves state-of-the-art distortion and perceptual metrics in 4 steps. P: PSNR, S: SSIM, L: LPIPS, F: FID.}
    \begin{subtable}[t]{\textwidth}
    \footnotesize
    \begin{tabularx}{\linewidth}{@{}l *{12}{>{\centering\arraybackslash}X} >{\centering\arraybackslash}X @{}}
    \toprule
    \multirow{2}{*}{Method} 
    & \multicolumn{4}{c}{$\sigma=15$} 
    & \multicolumn{4}{c}{$\sigma=25$} 
    & \multicolumn{4}{c}{$\sigma=50$} 
    \\
    \cmidrule(lr){2-5} \cmidrule(lr){6-9} \cmidrule(lr){10-13}
    & P$\uparrow$ & S$\uparrow$ & L$\downarrow$ & F$\downarrow$
    & P$\uparrow$ & S$\uparrow$ & L$\downarrow$ & F$\downarrow$
    & P$\uparrow$ & S$\uparrow$ & L$\downarrow$ & F$\downarrow$
    \\
    \midrule
    DnCNN 
    & 33.45 & .9035 & .068 & 37.14 
    & 31.52 & .8692 & .101 & 59.16  
    & 28.62 & .7986 & .173 & 107.31
    \\
    FFDNet 
    & 34.66 & \underline{.9216} & .065 & 39.37 
    & 32.36 & \underline{.8861} & .103 & 63.84 
    & \underline{29.19} & \underline{.8149} & .183 & 118.38
    \\
    IR-SDE 
    & \underline{34.80} & .9188 & \underline{.036} & \underline{22.03} 
    & \underline{32.39} & .8791 & \underline{.055} & \underline{34.66} 
    & 29.03 & .7911 & \underline{.091} & \underline{63.84}
    \\
    Baseline 
    & 33.51 & .8978 & .089 & 43.90 
    & 31.79 & .8697 & .122 & 66.47 
    & 29.15 & .8122 & .160 & 93.68
    \\
    ours-1
    & \textbf{35.40} & .9322 & .048 & 28.27
    & \textbf{33.15} & .9031 & .075 & 44.34
    & \textbf{30.13} & .8467 & .130 & 79.05
    \\
    ours-2
    & 35.37 & \textbf{.9324} & .039 & 21.89 
    & 33.11 & \textbf{.9033} & .062 & 33.18 
    & 30.08 & \textbf{.8470} & .110 & 63.06 
    \\
    ours-3
    & 35.29 & .9319 & .036 & 20.11 
    & 33.06 & .9021 & .057 & 30.15 
    & 29.98 & .8461 & .105 & 57.43 
    \\
    \rowcolor{gray!20}
    ours-4
    & 35.22 & .9310 & .033 & 16.97 
    & 32.93 & .9001 & .049 & 26.17 
    & 29.89 & .8439 & .100 & 52.64 
    \\
    ours-5
    & 35.18 & .9304 & .032 & 16.60 
    & 32.91 & .9001 & .050 & 26.00 
    & 29.81 & .8419 & .095 & 53.72 
    \\
    ours-10
    & 35.02 & .9278 & .027 & 14.45 
    & 32.68 & .8954 & .043 & 23.36 
    & 29.58 & .8343 & .087 & 49.86 
    \\
    ours-20
    & 34.88 & .9254 & \textbf{.025} & \textbf{13.39} 
    & 32.51 & .8906 & \textbf{.039} & \textbf{22.15} 
    & 29.39 & .8264 & \textbf{.082} & \textbf{48.22} 
    \\
    \bottomrule
    \end{tabularx}
    \end{subtable}

    \label{table:app-multi-step-noise-level-M}
\end{minipage}
\end{table*}

\begin{table*}[!htb]
\begin{minipage}{\linewidth}
    \centering
    \caption{Quantitative results of image denoising on Kodak24~\cite{kodak} test set.}
    \begin{subtable}[t]{\textwidth}
    \footnotesize
    \begin{tabularx}{\linewidth}{@{}l *{12}{>{\centering\arraybackslash}X} >{\centering\arraybackslash}X @{}}
    \toprule
    \multirow{2}{*}{Method} 
    & \multicolumn{4}{c}{$\sigma=15$} 
    & \multicolumn{4}{c}{$\sigma=25$} 
    & \multicolumn{4}{c}{$\sigma=50$} 
    \\
    \cmidrule(lr){2-5} \cmidrule(lr){6-9} \cmidrule(lr){10-13}
    & P$\uparrow$ & S$\uparrow$ & L$\downarrow$ & F$\downarrow$
    & P$\uparrow$ & S$\uparrow$ & L$\downarrow$ & F$\downarrow$
    & P$\uparrow$ & S$\uparrow$ & L$\downarrow$ & F$\downarrow$
    \\
    \midrule
    DnCNN 
    & 34.48 & .9189 & .083 & 21.71 
    & 32.02 & .8763 & .129 & 41.96 
    & 28.83 & .7908 & .229 & 83.27
    \\
    FFDNet 
    & 34.63 & \underline{.9215} & .085 & 21.57 
    & 32.13 & \underline{.8779} & .140 & 44.57 
    & \underline{28.98} & \underline{.7942} & .255 & 89.69
    \\
    IR-SDE 
    & \underline{34.64} & .9184 & \underline{.050} & \underline{13.74} 
    & 32.14 & .8739 & \underline{.078} & \underline{21.47} 
    & 28.75 & .7746 & \underline{.134} & \underline{45.96}
    \\
    Baseline 
    & 33.92 & .9090 & .110 & 24.52 
    & \underline{32.73} & .8666 & .161 & 45.81 
    & 28.89 & .7904 & .223 & 66.01
    \\
    ours-1
    & \textbf{35.31} & .9290 & .071 & 12.22 
    & \textbf{32.90} & .8919 & .107 & 21.66 
    & \textbf{29.90} & .8212 & .183 & 40.88 
    \\
    ours-2
    & 35.27 & \textbf{.9295} & .058 & 9.24 
    & 32.84 & \textbf{.8927} & .088 & 17.52 
    & 29.84 & \textbf{.8234} & .151 & 33.70
    \\
    ours-3
    & 35.19 & .9291 & .054 & 8.45  
    & 32.74 & .8923 & .084 & 16.44 
    & 29.72 & .8233 & .147 & 31.61  
    \\
    
    \rowcolor{gray!20}
    ours-4
    & 35.12 & .9283 & .049 & 7.21
    & 32.65 & .8909 & .078 & 15.25 
    & 29.62 & .8212 & .138 & 29.31  
    \\
    ours-5
    & 35.07 & .9276 & .047 & 7.07 
    & 32.58 & .8897 & .075 & 15.35 
    & 29.53 & .8190 & .131 & 29.45 
    \\
    ours-10
    & 34.91 & .9252 & .040 & 6.43 
    & 32.38 & .8854 & .067 & 14.51 
    & 29.26 & .8099 & .119 & \textbf{28.30}  
    \\
    ours-20
    & 34.80 & .9229 & \textbf{.037} & \textbf{6.35}  
    & 32.23 & .8811 & \textbf{.061} & \textbf{14.35} 
    & 29.04 & .7995 & \textbf{.116} & 28.44 
    \\
    \bottomrule
    \end{tabularx}
    \end{subtable}

    \label{table:app-multi-step-noise-level-K}
\end{minipage}
\end{table*}

\begin{table*}[!htb]
\begin{minipage}{\linewidth}
    \centering
    \caption{Quantitative results of image denoising on CBSD68~\cite{CBSD68} test set.}
    \begin{subtable}[t]{\textwidth}
    \footnotesize
    \begin{tabularx}{\linewidth}{@{}l *{12}{>{\centering\arraybackslash}X} >{\centering\arraybackslash}X @{}}
    \toprule
    \multirow{2}{*}{Method} 
    & \multicolumn{4}{c}{$\sigma=15$} 
    & \multicolumn{4}{c}{$\sigma=25$} 
    & \multicolumn{4}{c}{$\sigma=50$} 
    \\
    \cmidrule(lr){2-5} \cmidrule(lr){6-9} \cmidrule(lr){10-13}
    & P$\uparrow$ & S$\uparrow$ & L$\downarrow$ & F$\downarrow$
    & P$\uparrow$ & S$\uparrow$ & L$\downarrow$ & F$\downarrow$
    & P$\uparrow$ & S$\uparrow$ & L$\downarrow$ & F$\downarrow$
    \\
    \midrule
    DnCNN 
    & \underline{33.90} & \underline{.9289} & .063 & 25.59 
    & \underline{31.24} & \underline{.8830} & .109 & 43.51 
    & 27.95 & \underline{.7896} & .210 & 84.56
    \\
    FFDNet 
    & 33.88 & .9290 & .065 & 27.24 
    & 31.22 & .8821 & .121 & 49.64 
    & \underline{27.97} & .7887 & .244 & 98.76 
    \\
    IR-SDE 
    & 33.80 & .9251 & \underline{.042} & \underline{16.71} 
    & 31.14 & .8777 & \underline{.074} & \underline{28.71} 
    & 27.59 & .7733 & \underline{.138} & \underline{50.46}
    \\
    Baseline 
    & 33.02 & .9139 & .098 & 31.99 
    & 30.74 & .8661 & .162 & 56.64 
    & 27.84 & .7827 & .232 & 78.51
    \\
    ours-1
    & \textbf{34.31} & .9343 & .057 & 21.84
    & \textbf{31.70} & .8927 & .096 & 36.55
    & \textbf{28.52} & .8103 & .180 & 72.08 
    \\
    ours-2
    & 34.26 & \textbf{.9348} & .046 & 17.00
    & 31.63 & \textbf{.8938} & .078 & 28.64
    & 28.44 & \textbf{.8132} & .147 & 56.19 
    \\
    ours-3
    & 34.23 & .9344 & .043 & 15.74
    & 31.60 & .8930 & .071 & 26.25
    & 28.38 & .8117 & .132 & 51.12 
    \\
    \rowcolor{gray!20}
    ours-4
    & 34.11 & .9334 & .038 & 13.83 
    & 31.44 & .8922 & .068 & 23.71 
    & 28.21 & .8091 & .122 & 47.22 
    \\
    ours-5
    & 34.05 & .9329 & .038 & 13.73 
    & 31.38 & .8910 & .065 & 23.88
    & 28.12 & .8090 & .127 & 48.96 
    \\
    ours-10
    & 33.91 & .9308 & .034 & 12.61 
    & 31.20 & .8871 & .059 & 22.00 
    & 27.86 & .8006 & .118 & 44.35 
    \\
    ours-20
    & 33.81 & .9290 & \textbf{.033} & \textbf{12.19}
    & 31.06 & .8831 & \textbf{.056} & \textbf{21.27}
    & 27.65 & .7911 & \textbf{.114} & \textbf{42.49}  
    \\
    \bottomrule
    \end{tabularx}
    \end{subtable}

    \label{table:app-multi-step-noise-level-C}
\end{minipage}
\end{table*}

\section{Image generation}
For image generation on the CIFAR10 (32 × 32) dataset~\cite{cifar10}, we utilize the same U-net structure as DDIM~\cite{20DDIM}. We set the training patch-size to be $32\times 32$ and use a batch size of $128$. We used the Adam~\cite{adam} optimizer with parameters $\beta_1 = 0.9$. The total training steps are fixed to $600$ thousand and the initial learning rate set to $2e^{-4}$. ALL methods employ a linear schedule with $T = 1000$, we use the Frechet Inception Distance (FID) as the quantitative metric. The results demonstrate that our proposed cumulative velocity field significantly accelerates the convergence of velocity learning in rectified flow, while achieving performance comparable to DDIM, validating the effectiveness of our method.

\begin{figure*}[!htbp]
\centering
\setlength{\abovecaptionskip}{0.1cm}
\includegraphics[width=1.0\linewidth]{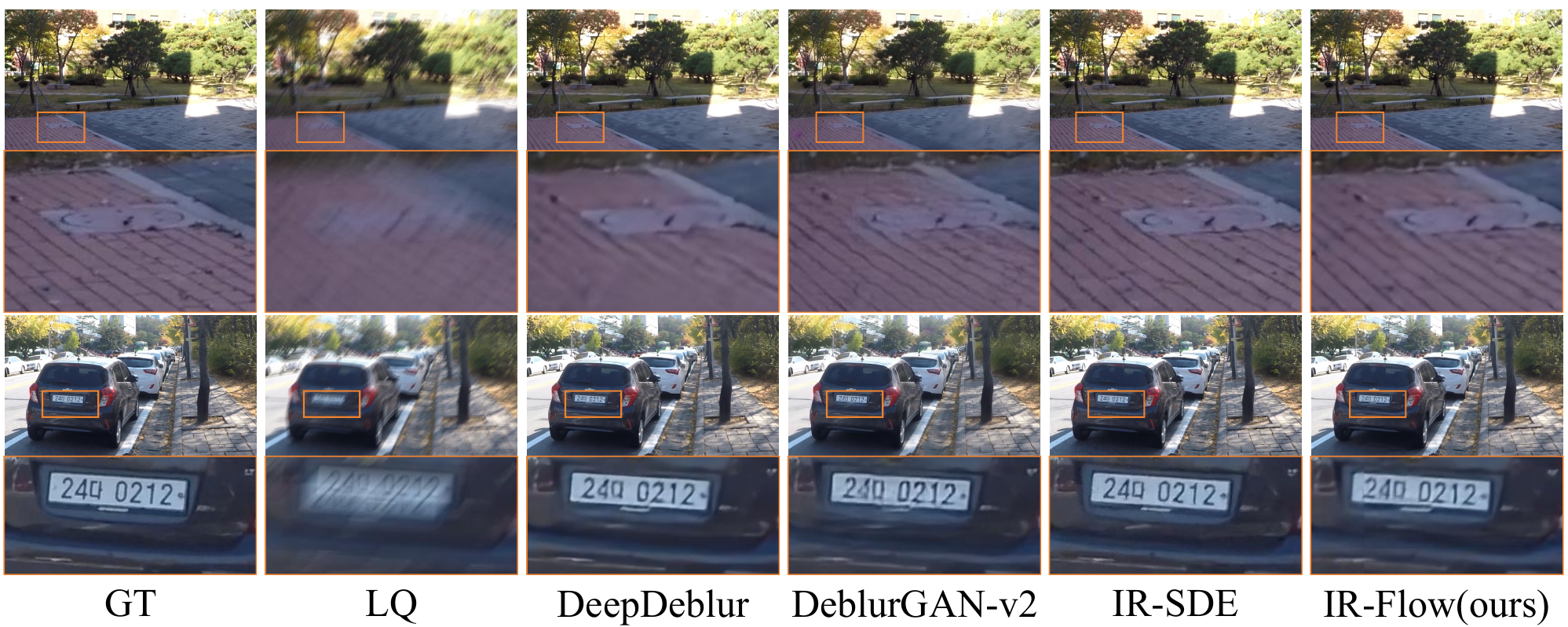}
\vspace*{-1em}
\caption{Visual results of the model on the real world deblurring GoPro dataset.}
\label{fig:GoPro}
\end{figure*}

\section{More visual results}
\label{sec:app-More-vision}

We first experiment on single image super-resolution, which is a fundamental and challenging task in computer vision. Our IR-Flow is trained and evaluated on the DIV2K~\cite{DIV2K} dataset, which contains 800 training and 100 testing images. As an additional preprocessing step, all the low-resolution images are bicubicly re-scaled to be of the same size as the corresponding high-resolution images. 

Image dehazing is often an important prerequisite for improving the robustness of other high-level vision tasks. Our IR-Flow is trained and evaluated on the RESIDE-6k\cite{hazedata} dataset, which includes a diverse mixture of indoor and outdoor scenes with 6,000 training and 1,000 testing images.

To further validate the effectiveness of our method in real-world scenarios, we conducted comparative experiments on real-world datasets SIDD~\cite{SIDD} for noisy images. 
 
We evaluate the deblurring performance of IR-Flow on the public GoPro dataset~\cite{Gopro} which contains 2103 image pairs for training and 1111 image pairs for testing. Compared with other synthetic blurry images from blur kernels, the GoPro dataset contains more realistic blur and is much more complex. To verify the generalization of our method, we perform deblurring training on NAFNet~\cite{NAFNet}.
Moreover, we compare our methods with several state-of-the-art deburring approaches such as DeepDeblur~\cite{Gopro}, DeblurGAN-v2~\cite{deblur14-DeblurGANv2}, DiffUIR~\cite{10377629}, IR-SDE~\cite{IRSDE}. The comparative experimental results are shown in Table~\ref{table:GoProD} and Figure~\ref{fig:GoPro}.

\section{Limitations and Future Work.}
Although our method focuses on image-based information learning, it lacks detailed textual guidance for severely degraded or occluded regions during restoration. In future work, we plan to extend IR-Flow toward joint text–image learning, enabling unified semantic guidance for a broader range of image restoration tasks. Moreover, since the current framework relies on supervised learning, its applicability to real-world scenarios is limited by the scarcity of paired data; thus, we will explore self-supervised learning strategies to enhance its practicality. In addition, we aim to establish a flexible framework that can seamlessly switch between SDE and ODE-based formulations, further improving the adaptability and efficiency of our approach.

\section{Societal Impact.}
This work presents societal benefits and risks inherent to image restoration. Benefits include enhanced smartphone photography and medical imaging for improved diagnostics. Conversely, risks involve privacy breaches from shared media and malicious image manipulation (e.g., deepfakes). We argue the societal advantages outweigh risks when supported by safeguards like digital watermarking and content authentication. To ensure responsible deployment, we advocate establishing ethical frameworks and technical countermeasures against misuse.

\begin{figure*}[!htb]
  \begin{center}
    \includegraphics[width=1.0\linewidth]{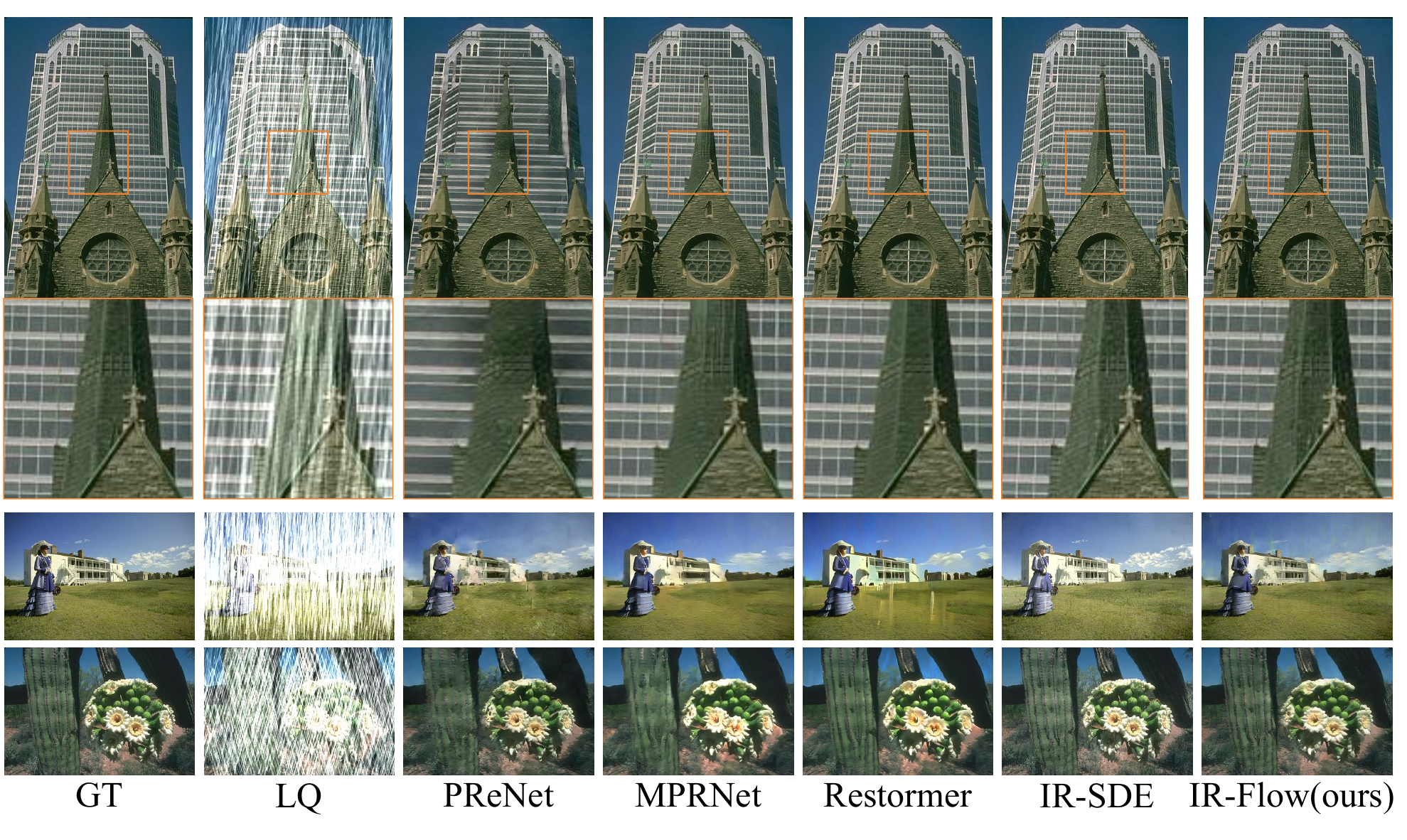}
  \end{center}
  \caption{More visual results on Rainy dataset.}
  \label{fig:app-rainHZ}
\end{figure*}

\begin{figure*}[!htb]
  \begin{center}
    \includegraphics[width=1.0\linewidth]{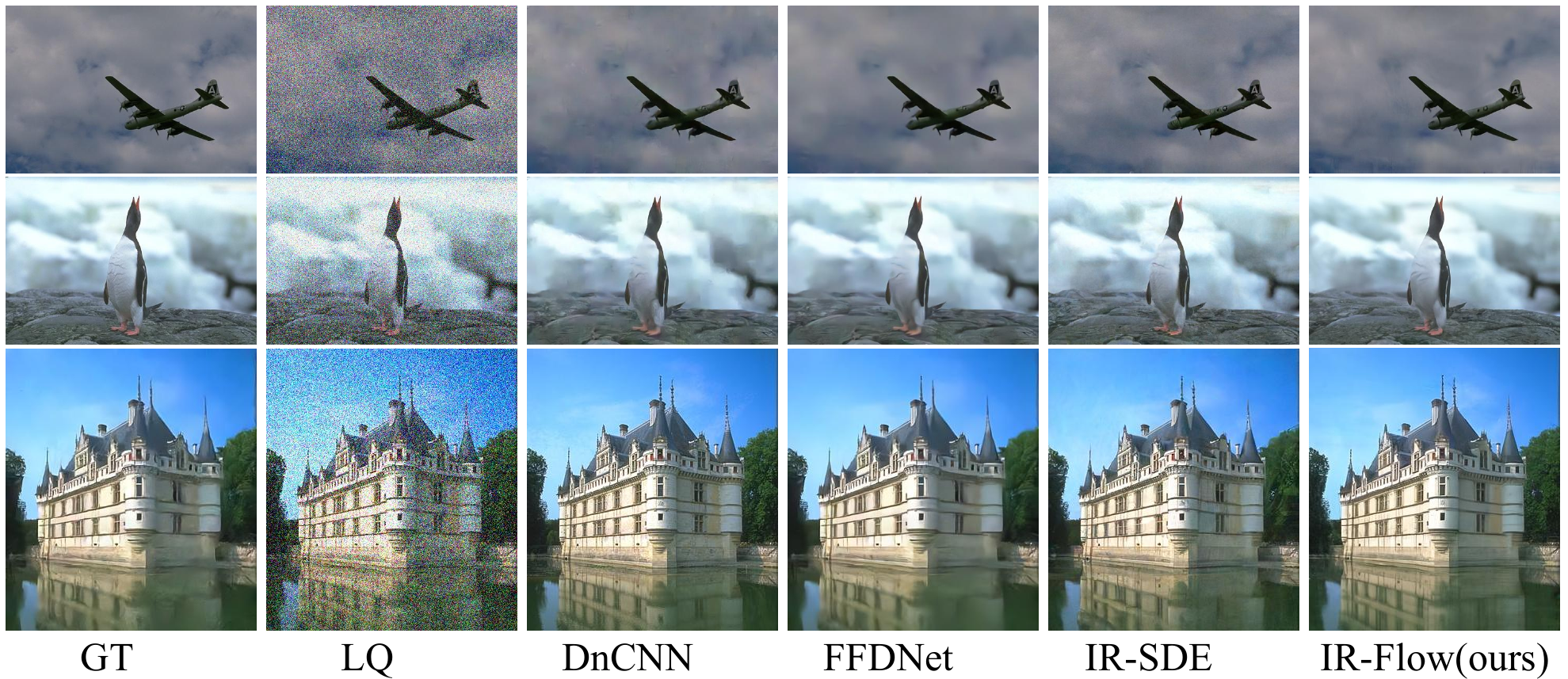}
  \end{center}
  \caption{More visual results on Noisy dataset.}
  \label{fig:app-noise}
\end{figure*}

\begin{figure*}[!htb]
  \begin{center}
    \includegraphics[width=1.0\linewidth]{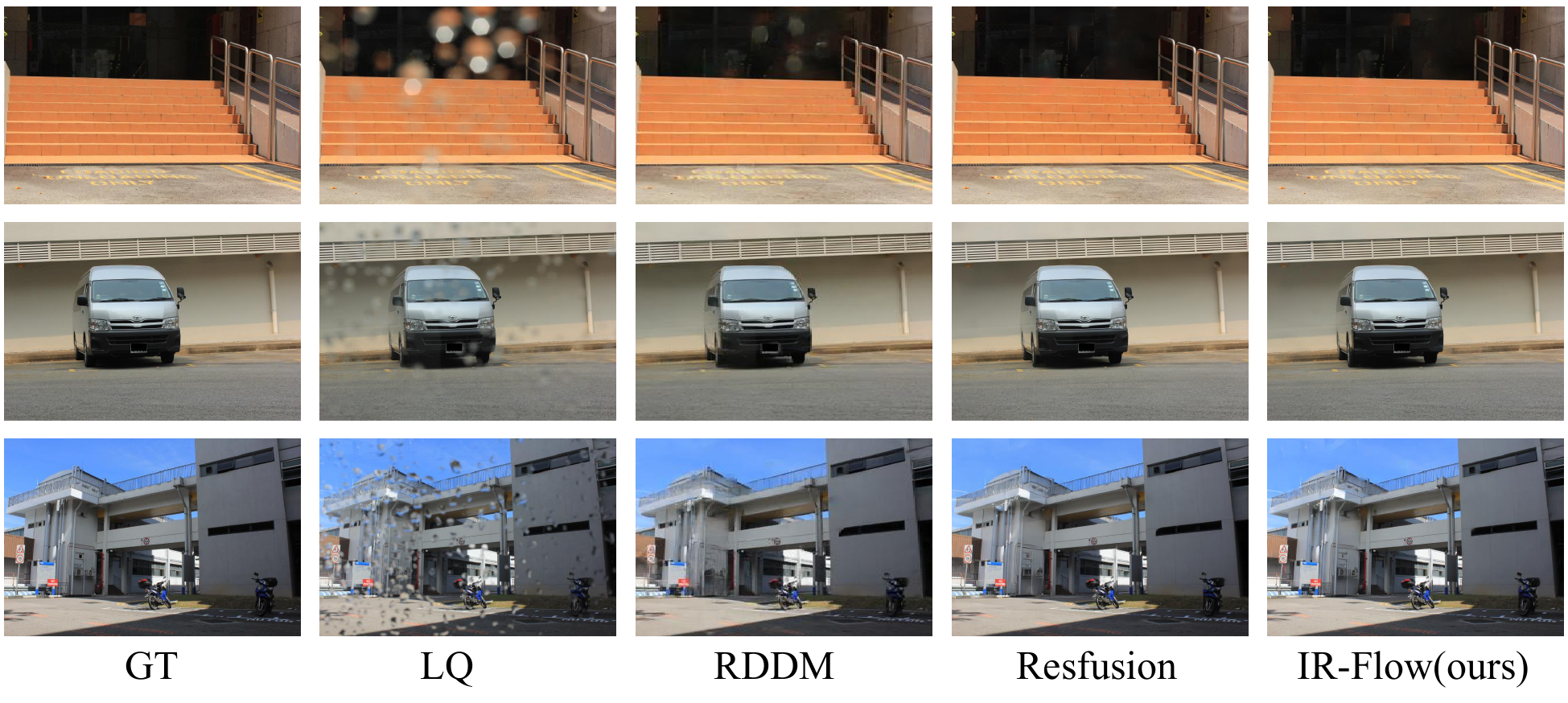}
  \end{center}
  \caption{Qualitative comparison results on Raindrop dataset.}
  \label{fig:app-rainDrop}
\end{figure*}

\begin{figure*}[!htb]
  \begin{center}
    \includegraphics[width=1.0\linewidth]{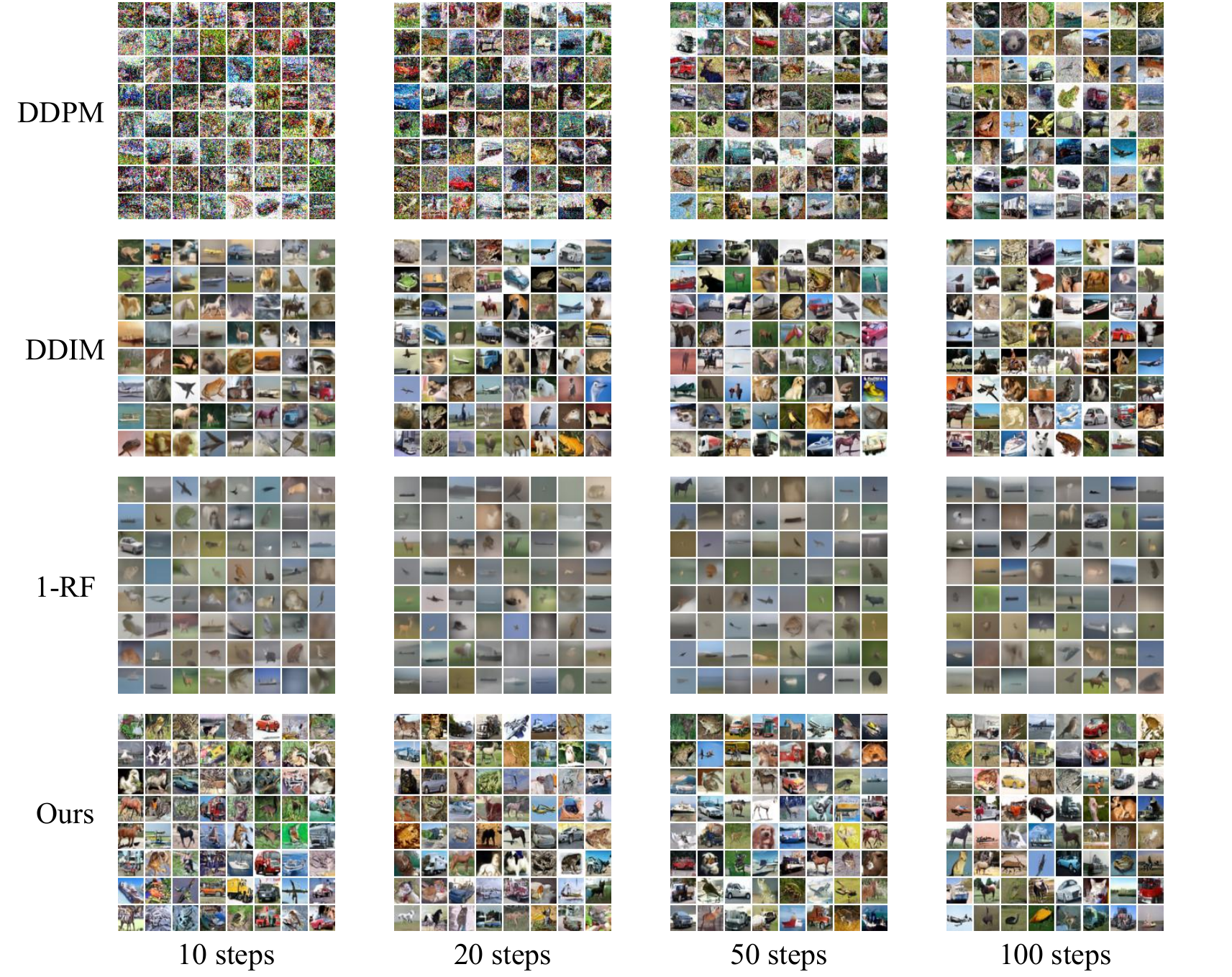}
  \end{center}
  \caption{More visual results between 1-Rectified Flow standard velocity and our velocity on CIFAR10 (32 × 32) dataset.}
  \label{fig:app-cifar}
\end{figure*}

\end{document}